\documentclass{article}

\usepackage{arxiv}

\usepackage{hyperref}
\hypersetup{hidelinks,
	backref=true,
	pagebackref=true,
	hyperindex=true,
	colorlinks=true,
	breaklinks=true,
	linkcolor= blue,
	urlcolor= blue,
	citecolor= blue,
	anchorcolor= blue,
}

\usepackage[numbers]{natbib}

\usepackage{breakcites}
\usepackage[ruled, vlined, linesnumbered]{algorithm2e}

\SetCommentSty{mycommfont}

\usepackage{amsthm}
\usepackage{amssymb}
\usepackage{amsmath,empheq, amsfonts}
\usepackage{bm}

\usepackage{graphicx}
\usepackage{svg}
\usepackage{tabu}
\usepackage{subcaption}
\usepackage{adjustbox}
\usepackage{placeins}
\usepackage{bbm}
\usepackage{empheq}
\usepackage{mathrsfs}
\usepackage{microtype}
\usepackage{booktabs}
\usepackage{empheq}
\usepackage{makecell}
\usepackage{multirow}

\usepackage[capitalize]{cleveref}

\newcommand{\tabfigure}[2]{\raisebox{-.5\height}{\includegraphics[#1]{#2}}}

\DeclareMathOperator{\diag}{diag}
\DeclareMathOperator{\tr}{Tr}

\DeclareMathOperator*{\argmin}{argmin}

\newcommand{\dimHx}{\mathrm{dim}(\mathscr{H}_{x})}
\newcommand{\dimHy}{\mathrm{dim}(\mathscr{H}_{y})}

\newcommand{\Hx}{\mathscr{H}_{x}}
\newcommand{\Hy}{\mathscr{H}_{y}}

\newcommand{\xoverbrace}[2][\vphantom{\sum_i^P}]{\overbrace{#1#2}}

	\title{Multi-view Kernel PCA for Time series Forecasting}

  \author{Arun Pandey \\
  Department of Electrical Engineering\\
  ESAT-STADIUS, KU Leuven\\
  Kasteelpark Arenberg 10, B-3001 Leuven, Belgium \\
  \texttt{arun.pandey@esat.kuleuven.be} \And Hannes De Meulemeester \\
  Department of Electrical Engineering\\
  ESAT-STADIUS, KU Leuven\\
  Kasteelpark Arenberg 10, B-3001 Leuven, Belgium\\
  \texttt{hannes.demeulemeester@esat.kuleuven.be}  \And \\
  \And Bart De Moor  \\
  Department of Electrical Engineering\\
  ESAT-STADIUS, KU Leuven\\
  Kasteelpark Arenberg 10, B-3001 Leuven, Belgium\\
  \texttt{bart.demoor@kuleuven.be}\And Johan A.K. Suykens \\
  Department of Electrical Engineering\\
  ESAT-STADIUS, KU Leuven\\
  Kasteelpark Arenberg 10, B-3001 Leuven, Belgium\\
  \texttt{johan.suykens@kuleuven.be}}

\begin{document}
\maketitle

	\begin{abstract}
		In this paper, we propose a kernel principal component analysis model for multi-variate time series
		forecasting, where the training and prediction schemes are derived from the multi-view formulation of Restricted Kernel Machines.
        The training problem is simply an eigenvalue decomposition of the summation of two kernel matrices corresponding to the views of the input and output data. When a linear kernel is used for the output view, it is shown that the forecasting equation takes the form of kernel ridge regression.
		When that kernel is non-linear, a pre-image problem has to be solved to forecast a point in the input space.
		We evaluate the model on
		several standard time series datasets, perform ablation studies, benchmark with closely related models and discuss its results.
	\end{abstract}

\section{Introduction}
Kernel methods have seen great success in many applications with very-high dimensional data but low number of samples, and are therefore one of the most popular non-parametric models.
In critical machine learning applications, kernel methods are preferred due to their strong theoretical foundation in learning theory \citep{vapnik95, Scholkopf2001, suykens_least_2002, gp_machine_learning_Rasmussen06}.
Kernel methods map the data into a high-dimensional (possibly infinite) feature space by using the kernel trick.
This kernel trick allows for natural, non-linear extensions to the traditional linear methods in terms of a dual representation using a suitable kernel function.
This led to numerous popular methods such as kernel principal component analysis \citep{kpca}, kernel fisher discriminant analysis \citep{kfda} and the least-squares support vector machine~\citep{vapnik95}.

However, when it comes to learning large-scale problems, kernel methods fall behind deep learning techniques due to their time and memory complexity.
This also holds in the time series analysis and forecasting domain, which has recently been dominated by specialized deep neural network models \citep{Oreshkin2020N-BEATS,Temporal_Fusion_Transformers_2021,fedformer_zhou22, nhits_2022_arxiv}.

Attempts have been made to combine kernel and deep learning methods \citep{Kernel_Methods_for_Deep_Learning_2009, Deep_Kernel_Learning_wilson16}, especially for specific  cases such as with deep gaussian processes \citep{deep_gaussian_processes} and multi-layer support vector machines \citep{multi_layer_svm}.
Recently, a new unifying framework, named Restricted Kernel Machines (RKM), was proposed \citep{suykensdeep2017} that attempts to bridge kernel methods with deep learning. The Lagrangian function of the Least-Squares Support Vector Machine (LS-SVM) is similar to the energy function of Restricted Boltzmann Machine (RBMs), thereby drawing link between kernel methods and RBMs; hence the name Restricted Kernel Machines.

\textbf{Contribution}:
In this work, we propose a novel kernel autoregressive
time series forecasting model based on the RKM framework, where the training problem is
the eigen-decomposition of two kernel matrices.
Additionally, we use the same objective function to derive a novel prediction scheme to recursively
forecast several steps ahead in the future. Further, we experimentally compare the model with several closely related models on multiple publically available time series datasets.

\section{Related Works}
\textbf{Energy-based methods}:
Energy-based models encode dependencies between visible and latent variables using the so-called energy function. Such a function is used to model joint-probability distribution over visible and latent variables of the system. Such a model is trained by finding a configuration with the lowest total energy.
Energy based models have the advantage over probabilistic methods that they do not require proper normalization, which in turn allows for more freedom when designing a model.
One of the most well known energy-based models is the Restricted Boltzmann Machine~\citep{salakhutdinov_restricted}.
Among other applications, this model has been extended for time series modeling~\citep{sutskeverTemporalRestricted, sutskeverRecurrentTemporalRestricted}.
For an excellent review of Energy-based Models and Restricted Boltzmann Machines for
time series
modeling, see~\cite{osogamiBoltzmannMachinesEnergybased2019, osogamiBoltzmannMachinesTimeseries2019}.

\textbf{Kernel methods for time series}:
Kernel methods have long been a staple in the time series analysis toolkit.
Thanks to Mercer's theorem \citep{mercer_james_functions}, these methods can
implicitly work in very high dimensional spaces and can often allow for
complicated non-linear problems to be turned into convex optimization problems.

In the case of time series forecasting, one can change the traditional kernel ridge regression formulation into an
online variant where only a number of data points are available at each time
step. The solution to this problem can then be iteratively updated for each new
time step. This technique is known as Kernel Recursive Least-Squares~\citep{KRSL} (KRLS).
One disadvantage to this method is that the model increases in
size for each new data point that gets introduced. Various solutions for this
have been proposed that attempt to limit the final model size~\citep{sw_krls, fd_krls}.
Alternatively, a bayesian formulation for KRLS exists which also
includes a forgetting mechanism~\citep{krls_tracker}.

Support vector machines \citep{SVM_timeseries, suykens_least_2002} are a
particular popular type of kernel machine since training an SVM corresponds with
solving a convex optimization problem with sparse solutions. The standard,
static, regression variants of support vector machines can be used in
feedforward fashion for time series forecasting. Alternatively, explicitly
recursive extensions have been formulated \citep{recurrent_lssvm}. The
disadvantage is that the resulting optimization problems become non-convex.

Gaussian processes~\citep{GP_regression} are the extension of kernel methods to
the probabilistic setting. Unlike support vector machines, gaussian processes
are not sparse, which can be problematic for large scale problems.
Various
solutions have been proposed to reduce the effective number of data points and
train on large data sets, such as by using inducing variables~\citep{sparse_gps}
or by performing the computations on subsets of the data~\citep{
	deisenroth2015distributed}. \cite{gp_lvm} showed that a gaussian process could
be formulated as a latent variable model. This allowed for hierarchical, deep,
gaussian processes~\citep{deep_gaussian_processes}.

\subsection{Restricted Kernel Machines}
Restricted Kernel Machines (RKM) \citep{suykensdeep2017} is a general kernel
methods framework characterized by visible and hidden units. 
Using the Fenchel-Young conjugate for quadratic functions, RKMs make the connection between
 Least-Squares Support Vector Machines
\citep{suykens_least_2002} and Restricted Boltzmann Machines
\citep{salakhutdinov_restricted}.
The workhorse of restricted kernel machines is
kernel PCA, of which the energy function can be defined in the RKM framework
using visible ($\bm{v}_i$) and latent ($\bm{h}_i$) units as follows:
\begin{empheq}{align}\label{obj:general_rkm}
	J(\bm{W},\bm{h}_i) =
	\sum_{i=1}^{n}
	\langle \bm{v}_{i}, \bm{W} \bm{h}_i \rangle_{\mathscr{H}}
	+ \dfrac{1}{2} \bm{h}_{i}^\top \bm{\Lambda}  \bm{h}_i
	+ \dfrac{1}{2}  \tr (\bm{W}^{\top} \bm{W})
\end{empheq}
The stationary points of the energy function are given by:
\begin{align*}
	\label{eq:general_rkm_saddle}
	\frac{\partial{J}}{\partial{\bm{h_i}}} & = 0  \implies \bm{W^\top}\bm{v}_{i}
	= \bm{\Lambda} \bm{h_i}, \quad \forall i                                             \\
	\frac{\partial{J}}{\partial{\bm{W}}}   & = 0  \implies \bm{W} = \sum_{i=1}^{n}
	\bm{v}_{i} \bm{h}_{i} ^{\top}.
\end{align*}
Eliminating the $\bm{W}$ from above yields the foloowing eigenvalue problem:
\begin{equation}\label{eq:general_rkm_KPCA}
	\bm{K}\bm{H}^\top=\bm{H}^\top \bm{\Lambda}
\end{equation}
with $\bm{H} =\big[\bm{h}_1 \ldots \bm{h}_n \big]\in \mathbb{R}^{s\times N} $
and $ \bm{\Lambda} =\diag(\lambda_i)\in\mathbb{R}^{s\times s} $ where generally
$s \leq N$, indicating the number of components. Training an RKM model consists
of solving an eigenvalue problem. This is in contrast with restricted boltzmann
machines where the model is trained by approximating the gradient of the
log-likelihood of $\bm{W}$ by using methods such as contrastive divergence
\citep{rbm_constrastive_divergence}.

The objective function $J(\bm{W},\bm{h}_i) $ in~\cref{obj:general_rkm} forms
the basis for more specific RKM models which can be used in downstream tasks.
Additionally, a key advantage of characterizing kernel methods by visible and
hidden units is that deeper, multi-layer, models can be developed similarly to
deep boltzmann machines~\citep{salakhutdinov_deep}.

Classification and regression can be performed by introducing the primal
formulation of LS-SVM into the objective function~\citep{suykensdeep2017}. This
is motivated by the fact that kernel PCA corresponds to a one-class LS-SVM
problem with zero target value around which the variance is maximized
\citep{suykens2003support}. Classification using RKMs has been extended
to multi-view classification using a tensor-based representation of the methods
\citep{HOUTHUYS202154}. RKMs have been shown to be powerful feature extractors
and can be used for disentangled representation learning~\citep{TONIN2021661,
	WinantLatentSpace}.
This, in turn, allows them to be used in outlier and
out-of-distribution detection tasks \citep{ToninOOD}.
Finally, the latent space
of an RKM can be sampled from to generate new samples from the learned
distribution \citep{joachim, robust2020, GENRKM, strkm}.

One aspect that was missing, until now, was a general time series model situated
in the RKM setting. This work attempts to fill this gap by introducing a
framework with autoregressive RKMs. This idea was briefly introduced in~\citep{rrkm_esann} where an RKM model with autoregression in the latent space
was discussed. In the next section, we generalize this method and introduce a
complete framework for time series analysis using restricted kernel machines.

\section{Proposed Model}

Our  objective is to capture the dynamics of the multi-variate time series
$\{\bm{{x}}_{i}\}_{i=1} ^{n}, \bm{{x}}_{i} \in \mathbb{R}^{d}$
containing $n$ time
steps.
Given the lag $p \in \mathbb{Z}^{+}$, we transform the given time series such
that the input vector is a vector of lagged values and the corresponding output
vector is one-step ahead data-point.
Let the input data-vectors be
$
	\{[\bm{{x}}_i^{(1)} , \ldots , \bm{{x}}_{i-p}^{(1)},
	\bm{{x}}_i^{(d)},
	\ldots , \bm{{x}}_{i-p}^{(d)}]\}_{i=p+1}^{n-1} \subset \mathcal{X}
$
and the corresponding target values be
$ \{\bm{{x}}_{i+1}\}_{i=p+1}^{n-1} \subset \mathcal{Y} $. This model structure is employed to model the dynamics of the system which enables us to use static regression models for training.
However, for prediction, we employ the \emph{recursive} model such that the predictions are used as input for further predictions ${\bm{\hat{y}}} = f( \bm{\hat{y}}_{i-1}, \ldots ,  \bm{\hat{y}}_{i-p})$.

We define feature maps $\phi: \mathscr{X} \rightarrow \Hx$, $\psi: \mathscr{Y}
	\rightarrow \Hy$ with $\Hx, \Hy$ indicating the Reproducing Kernel Hilbert Space\footnote{Throughout our discussion,
we assume that the feature vectors are
centered in the feature-space \emph{i.e.} (possibly infinite dimensional)
$\bar{\phi}(\bm{x}) = \phi(\bm{x}) - \mu_{\phi}$ with $\mu_{\phi} =
	\mathbb{E}_{\bm{\xi}\mathcal{\sim X}}\left[\phi(\bm{\xi})\right]$. Using an
implicit formulation, it suffices to notice that
$\langle\bar{\phi}(\bm{x}),
	\bar{\phi}(\bm{y}) \rangle_{\mathscr{H}} = \langle\phi(\bm{x}) - \mu_{\phi},
	\phi(\bm{y}) - \mu_{\phi} \rangle_{\mathscr{H}} = \langle\phi(\bm{x}),
	\phi(\bm{y})\rangle_{\mathscr{H}} - \langle\mu_{\phi},
	\phi(\bm{y})\rangle_{\mathscr{H}} - \langle\phi(\bm{x}),
	\mu_{\phi})\rangle_{\mathscr{H}} + \langle\mu_{\phi},
	\mu_{\phi}\rangle_{\mathscr{H}} = k(\bm{x},\bm{y}) -
	\mathbb{E}_{\bm{\xi}\mathcal{\sim X}}\left[k(\bm{\xi},\bm{y})\right] -
	\mathbb{E}_{\bm{\zeta}\mathcal{\sim X}}\left[k(\bm{x},\bm{\zeta})\right] +
	\mathbb{E}_{\bm{\xi},\bm{\zeta}\mathcal{\sim
		X}}\left[k(\bm{\xi},\bm{\zeta})\right]$.
In practice, we can compute
statistics on $\mathcal{X}$.}
(RKHS; see~\citep{scholkopf1997kernel} for more
details).
Such a feature map could be constructed explicitly or implicitly via a
kernel function
$ k(\bm{x}_{i},\bm{x}_{j}) : \mathscr{X}^2 \rightarrow
	\mathbb{R} : \left(\bm{x}_{i},\bm{x}_{j} \right) \mapsto
	\langle\phi(\bm{x}_{i}), \phi(\bm{x}_{j})\rangle_{\Hx}$.
Further we define two linear operators\footnote{The linear operator
$\bm{U}_{x}$ is often referred to as a \emph{matrix} as it only exists
explicitly in the case of finite dimensional Hilbert spaces $\mathscr{H}$. It
then takes the form $\bm{U}_{x} \in \mathbb{R}^{\mathrm{dim}(\mathscr{H}_x)
	\times s}$.}
$\bm{U}_{x}:\mathbb{R}^s \rightarrow \Hx$ and
$\bm{U}_{y}:\mathbb{R}^s \rightarrow \Hy$, $s \leq n$ with their ad-joints
$\bm{U}_{x}^{\top}: \dimHx \rightarrow \mathbb{R}^{s}$, $\bm{U}_{y}^{\top}: \dimHy
	\rightarrow \mathbb{R}^{s}$.
Data points $\bm{x}_i$ and $\bm{y}_i$ will be associated to a latent variable
$\bm{h}_{t} \in \mathbb{R}^s$ through the pairing terms $\langle
	\phi\left(\bm{x}_i\right), \bm{U}_{x} \bm{h}_i \rangle_{\mathscr{H}_x}$ and
$\langle \psi\left(\bm{y}_i\right), \bm{U}_{y} \bm{h}_i \rangle_{\mathscr{H}_y}$.

\subsection{Training}
\begin{empheq}{align}
	\label{obj:dual_rrkm_training}
	J_{n}(\bm{U}_{x},\bm{U}_{y},\mathcal{H}_{n})
	= &\sum_{i=1}^{n} \big[
	- \xoverbrace{
		\langle \phi (\bm{x}_{i}), \bm{U}_{x}\bm{h}_i \rangle_{\Hx}
	}^{
	\text{1\textsuperscript{st} view}
	}
	- \xoverbrace{
		\langle \psi
		(\bm{y}_{i}), \bm{U}_{y} \bm{h}_i \rangle_{\Hy}
	}^{
	\text{2\textsuperscript{nd} view}
	}
	+\dfrac{1}{2} \bm{h}_{i}^\top \bm{\Lambda} \bm{h}_i \nonumber\\
	& \underbrace{
	+ \lVert
	\phi(\bm{x}_{i})\rVert_{\Hx}^{2}
	+ \lVert
	\psi(\bm{y}_{i})\rVert_{\Hy}^{2}\big]
	+ \dfrac{1}{2}
	\big[ \lVert \bm{U}_{x} \rVert^{2}_{\mathrm{F}}
	+ \lVert \bm{U}_{y} \rVert^{2}_{\mathrm{F}} \big]
	}_{\text{regularization}}
\end{empheq}
The first two terms
in the objective maximizes the pairing between the visible variables in
the feature-space $\left\{\phi(\bm{x}) : \bm{x} \in \mathscr{X} \right\}$,
$\left\{\psi(\bm{y}) : \bm{y} \in \mathscr{Y} \right\}$ and latent variables
$\left\{\bm{h} \in \mathbb{R}^s \right\}$.
In this way we map the points from
the two data-sources to a single point in the latent space.
The regularization terms and
constraints are meant to bound the objective.

\paragraph{Solving the objective}
Given the visible variables, characterizing the stationary points of
$J(\bm{U}_{x},\bm{U}_{y},\mathcal{H}_T|\mathcal{X}_T)$ in the pairing
linear operators and the  latent variables leads to the following equations for
$1  \leq i \leq n$:
\begin{empheq}[left=\empheqlbrace]{align}
	\dfrac{\partial J}{\partial
		\bm{h}_{i}} &
	=0       \Rightarrow
	\bm{\Lambda} \bm{h}_i =  \bm{U}_{x}^\top \phi(\bm{x}_i) +\bm{U}_{y}^\top
	\psi(\bm{y}_i) \label{eq:dJ_over_dh_i}\\
	\dfrac{\partial J}{\partial \bm{U}_{x}} &
	=0
	\Rightarrow                                          \bm{U}_{x} =
	\sum_{i=1}^{n}\phi(\bm{x}_{i}) \bm{h}_{i}^{\top}  \label{eq:dJ_over_dUx}\\
	\dfrac{\partial J}{\partial \bm{U}_{y}} &
	=0
	\Rightarrow \bm{U}_{y} = \sum_{i=1}^{n}\psi(\bm{y}_{i})  \bm{h}_{i}^{\top}.
	\label{eq:dJ_over_dUy}
\end{empheq}
Eliminating $\bm{U}_{x}, \bm{U}_{y}$ from the above equations and denoting $
	\bm{H} =\big[\bm{h}_1 \ldots \bm{h}_n \big]\in \mathbb{R}^{s\times n} $ and
$ \bm{\Lambda} =\diag(\lambda_i)\in\mathbb{R}^{s\times s} $, we get the
following eigen-value problem:
\begin{equation}
	\label{eq:KPCA}
	\Big[  \bm{K}(\mathcal{X}) +
		\bm{K}(\mathcal{Y})  \Big] \bm{H}^\top
	=\bm{H}^\top \bm{\Lambda},
\end{equation}
where the elements of $ \bm{K}(\mathcal{X})_{i,j} = k_{x}(\bm{x}_i, \bm{x}_j) $ and $
	\bm{K}(\mathcal{Y})_{i,j} = k_y(\bm{y}_i, \bm{y}_j) $.
Further,
$\bm{K}(\mathcal{X})$ and $\bm{K}(\mathcal{Y})$ are
Hermitian matrices and their sum is also Hermitian. \emph{Spectral theorem} guarantees
the existence of real eigenvalues when diagonalized by a unitary matrix.
Moreover these eigenvalues are non-negative, since these matrices are positive
semi-definite.

\paragraph{Remarks}
The proposed model is inspired by~\citep{rrkm_esann}, however there are three
important differences.
(i) The data in the proposed model is in the auto-regressive format.
This enables us to formulate the problem in the well-studied Non-linear
Auto-Regressive (NAR) model setting.
(ii) The second-view $ \bm{K}(\mathcal{Y}) $ is now a kernel matrix i.e. a
positive semi-definite matrix. This was not a constraint with the \emph{time-kernel} in~\cite{rrkm_esann}.
Moreover, $ \bm{K}(\mathcal{Y}) $ implicitly captures the weights on the
past latent-variables via a kernel function instead of hand-crafting it.
(iii) The prediction scheme in~\cite{rrkm_esann} used past latent variables
to predict the next latent variable. This was sufficient to simulate the
dynamics forward after the training-set, however there was no way to start
with \emph{any} given initial state vector.
In contrast, here we require the initial state vector from the input space to
predict the dynamics forward.

\subsection{Prediction}

The main goal is to forecast ahead of the training set i.e. find
$\left\{\bm{x}_{n+1},\ldots, \bm{x}_{n^{\prime}}\right\}$, $n^{\prime} > n$.
To do so, we now work in
$\mathcal{X}_{n^{\prime}} = \mathcal{X}_n \cup \left\{\bm{x}_{n+1},\ldots,\bm{x}_{n+n^{\prime}}\right\}$,
$\mathcal{H}_{n^{\prime}} =
	\mathcal{H}_n \cup \left\{\bm{h}_{n+1},\ldots,\bm{h}_{n+n^{\prime}}\right\}$ and consider
$\mathcal{A}_{n^{\prime}}$. This gives the following objective
\begin{align}
	\label{eq:obj_pred}
	J_{n^{\prime}}(\bm{U}_{x},\bm{U}_{y},\mathcal{H}_{n^{\prime}})
	= & \sum_{i=1}^{n^{\prime}} \big[-
		\langle \phi (\bm{x}_{i}), \bm{U}_{x}\bm{h}_i \rangle_{\Hx}
		- \langle \psi
		(\bm{y}_{i}), \bm{U}_{y} \bm{h}_i \rangle_{\Hy}
	+   \dfrac{1}{2} \bm{h}_{i}^\top \bm{\Lambda} \bm{h}_i \nonumber \\
	  & + \lVert \phi(\bm{x}_{i})\rVert_{\Hx}^{2}
		+ \lVert \phi(\bm{y}_{i})\rVert_{\Hy}^{2} \big]
	+ \dfrac{1}{2}
	\big[ \lVert \bm{U}_{x} \rVert^{2}_{\mathrm{F}}
	+ \lVert \bm{U}_{y} \rVert^{2}_{\mathrm{F}} \big]
\end{align}
Given the learned $\bm{U}_{x}, \bm{U}_{y}$ from the training, characterizing the stationary
points of $J_{n^{\prime}}\left(\mathcal{X}_{n^{\prime}},\mathcal{H}_{\prime}
	\left|\bm{U}_{x}, \bm{U}_{y}\right.\right)$ in terms of visible and latent variables gives for
$1 \leq i \leq n^{\prime}$:
\begin{empheq}[left=\empheqlbrace]{align}
	\dfrac{\partial J}{\partial
		\phi(\bm{x}_{i})}  &
	=0
	\Rightarrow \phi(\bm{x}_{i}) =  \bm{U}_{x}\bm{h}_{i}  \label{eq:x_star_Uh}
	\\
	\dfrac{\partial J}{\partial \phi(\bm{y}_{i})}  &
	=0 \Rightarrow \psi(\bm{y}_{i}) =
	\bm{U}_{y}\bm{h}_{i}  \label{eq:y_star_Uh} \\
	\dfrac{\partial J}{\partial \bm{h}_{i}}  &
	=0 \Rightarrow \bm{\Lambda} \bm{h}_{i} - \bm{U}_{y}^\top
	\psi(\bm{y}_{i}) = \bm{U}_{x}^\top \phi(\bm{x}_{i}).
	\label{eq:h_star_x_star_y_star}
\end{empheq}
Using the above equations, we can obtain an expression for
$\bm{h}_{i}$ with $\phi(\bm{x}_{i})$:
\begin{empheq}{equation}
	\bm{h}_{i} = \big(
	\bm{\Lambda} - \bm{U}_{y}^\top \bm{U}_{y}
	\big)^{-1}\bm{U}_{x}^{\top}\phi({\bm{x}_{i}})\label{eq:h_i_from_x}.
\end{empheq}
Substituting it back
in~\cref{eq:y_star_Uh}, we obtain an expression for output in the feature-space:
\begin{empheq}{align}
	\psi(\bm{y}_{i}) &
	= \bm{U}_{y}
	\big( \bm{\Lambda} - \bm{U}_{y}^\top \bm{U}_{y}
	\big)^{-1}\bm{U}_{x}^{\top}\phi(\bm{x}_{i})
	\label{eq:phi_y_hat}.
\end{empheq}

\textbf{Case I: Non-linear $k_x$, linear $k_y$}:
In this case,~\cref{eq:phi_y_hat} reduces to the following variant of kernel ridge regression
$
	\bm{y}_{i}  = \big(\sum_{j=1}^{n} \tilde{\bm{h}}_j k_x ( \bm{x}_j, \bm{x}_i ) \big) $, where $\tilde{\bm{h}} = \bm{U}_{y}
	\big( \bm{\Lambda} - \bm{U}_{y}^\top \bm{U}_{y}
	\big)^{-1} \bm{h}_j
$. The way in which those coefficients are obtained, differs between our method
and kernel ridge regression.

\textbf{Case II: Non-linear $k_x$, non-linear $k_y$}:
If the feature-map $ \psi(\cdot) $  is explicitly known and \emph{invertible},
then after the inversion, we have the desired output $\bm{y}_{i}$.
However typically in kernel methods, for instance when using Gaussian kernel, $
	\psi(\cdot)$ is unknown. As a consequence, the closed-form expression
in~\cref{eq:phi_y_hat} cannot be computed.
In such situations, we leverage the \emph{kernel-trick} to obtain the
distances between the implicit feature-vector and the feature-vectors of the
training-set.

\textbf{Kernel-trick}:
Left-multiplying~\cref{eq:phi_y_hat} with $\phi
	(\bm{y}_j)^\top$ and substituting for
$\bm{U}_{y}$ from~\cref{eq:dJ_over_dUy}, we
obtain $ \forall  j \in \{1, \ldots , n\} $:
\begin{empheq}{align*}
	k_y(\bm{y}_j, \bm{y}_{i})=
	\Big(\sum_{k=1}^{n}
	k_y(\bm{y}_j,\bm{y}_k)\bm{h}_{k}^{\top} \Big)
	\Big( \bm{\Lambda} -  \sum_{l,m = 1}^{n}
	\bm{h}_l k_y(\bm{y}_l,\bm{y}_m) \bm{h}_{m}^{\top}\Big)^{-1} \\
	\Big( \sum_{k=1}^{n}
	\bm{h}_k k(\bm{x}_k,\bm{x}_{i}) \Big).
\end{empheq}
The above could be reduced in the matrix form as follows:
\begin{empheq}{align}
	\bm{k}_{\bm{y}}(\bm{y}_{i})
	= {\bm K (\mathcal{Y})}\bm{H}^{\top}\Big(
	\bm{\Lambda} -  \bm{H} {\bm K(\mathcal{Y})} \bm{H}^{\top}\Big)^{-1} \Big(
	\bm{H}
	\bm{k}_{\bm{x}}(\bm{x}_{i}) \Big),
	\label{eq:k_y_y_i}
\end{empheq}
where
$\bm{k}_{\bm{y}}(\bm{y}_{i})= [{k}_{y}(\bm{y}_1,\bm{y}_{i}), \dots,
	{k}_{y}(\bm{y}_{n},\bm{y}_{i})]^\top$.
These distances in the feature space
can be used in various \emph{pre-image methods} to obtain the output $\bm{y}_{i}$.
In the next section,  we briefly
discuss the pre-image problem and some common methods.

\subsubsection{Pre-image Problem}

The advantage of using a kernel, $ k(\bm{y}_{t},\bm{y}_{j}  ) =
	\langle\phi(\bm{y}_{i}), \phi(\bm{y}_{j}) \rangle_{\mathcal{H}_y} $,  is that all computations
can be implicitly performed in feature space and the exact mapping $\psi:
	\mathcal{Y} \rightarrow \mathcal{H}_{y} $ is not required.
However, this gives rise to the pre-image problem: given a point $\Psi
	\in \mathcal{H}$, find $ \bm{y} \in \mathcal{Y}$ such that $\Psi =
	\psi(\bm{y})$.
This pre-image problem is known to be
ill-posed~\citep{mika_kernel_nodate} as the pre-image might not exist
\citep{Scholkopf2001} or, if it exists, it may not be unique.
Instead, an optimization problem is formulated to find
the approximate pre-image $\bm{y}^\star$:
\begin{equation}
	\underset{\bm{y}^\star}{\argmin }\Vert \Psi-\psi(\bm{y}^\star)\Vert^{2}
	\label{eq:preimage_optimization}
\end{equation}

While various techniques exist to approximately solve this
problem~\citep{bui_projection-free_2019,
	kwok_pre-image_2004-2,
	honeine_preimage_2011-1,weston_learning_2004},
we discuss two particular methods that can conveniently be incorporated into the
RKM framework.

\textbf{Kernel Smoother}:
A kernel smoother~\citep{joachim} is essentially a weighted-average of nearby data-points in the input space where both the weights and the nearby-points are given by the kernels in~\cref{eq:k_y_y_i}.
Let $n_r < N$  be the number of nearest-neighbors, then the approximate pre-image for time-step $i$ is given by:
\begin{equation*}
	\tilde{\bm{y}}_i = \sum_{j=1}^{n_r} \frac{
		{k}_{y}\left(\bm{y}_{j}, \bm{y}_i\right) \bm{y}_{j}}{\sum_{l=1}^{n_r}
		{k_y}\left(\bm{y}_{l}, \bm{y}_i\right)}.
	\label{eq:kernel_smoother}
\end{equation*}
With this method, the implicit assumption is that the forecasted point lives within the simplex of training data-points. 
Further, in the limit-case $n_r = 1$, only the nearest training point is chosen as the prediction. In our experiments, we treat $n_r$ as the hyperparameter.

\textbf{Kernel Ridge Regression}:
Alternatively, one can \emph{learn} the pre-image map using kernel ridge regression \citep{weston_learning_2004}.
Consider a specific case of \cref{eq:preimage_optimization} where the denoised point $\Psi = \bm{U}_y \bm{h}^\star$ (see~\cref{eq:y_star_Uh}) i.e.
$ \underset{\bm{x}^\star}{\argmin} \Vert \bm{U}_y \bm{h}^\star - \psi(\bm{y}^\star) \Vert^2 $.
Then one could formulate the following problem to learn the pre-image map $\Gamma \colon \mathcal{H} \mapsto \mathcal{Y}$ as follows:
\begin{equation*}
	\argmin_{\Gamma} \sum_{i=1}^{n} \ell (\bm{y}_i, \bm{U}_y \bm{h}_i) + \lambda \Omega (\Gamma).
\end{equation*}
Since $\mathcal{Y}$ is a vector space $\mathbb{R}^{d}$, we can consider the above as kernel ridge regression problem with kernel $k$. This gives the mapping $ \Gamma_j(\bm{U}_y \bm{h}_i) = \sum_{r=1}^{n} \alpha_r^j k(\bm{U}_y \bm{h}_r, \bm{U}_y \bm{h}_i) = \sum_{r=1}^{n} \alpha_r^j k(\bm{h}_r, \bm{h}_i), \forall j \in \{1, \ldots , n\}$ where the coefficient are obtained by
$\alpha = \left(\bm{K}^\top \bm{K} + \lambda
	\mathbb{I}\right)^{-1} \bm{K} \bm{Y}$.
\begin{algorithm}[ht]
	\DontPrintSemicolon
	\KwIn{Time series $\{\bm{{x}}_{i}\}_{i=1} ^{n}, \bm{{x}}_{i} \in \mathbb{R}^{d}$  }
	\KwOut{Optimal kernel parameters}

	\While{\textrm{validation error} $ \geq $ 0}{
	Init. kernels $ k_x , k_y $; lag $ p $  \tcp*{Sample from grid}

	$ \bm{X} \gets
		\{[\bm{{x}}_i^{(1)} , \ldots , \bm{{x}}_{i-p}^{(1)},
		\bm{{x}}_i^{(d)},
		\ldots , \bm{{x}}_{i-p}^{(d)}]\}_{i=p+1}^{n-1} $ \tcp*{AR format}

	$ \bm{Y} \gets \{\bm{{x}}_{i+1}\}_{i=p+1}^{n-1} $ \tcp*{One-step ahead}

	$\bm{K}(\mathcal{X})  \gets \bm{{X}}$, $\bm{K}(\mathcal{Y})
		\gets \bm{{Y}}$   \tcp*{kernel matrices}

	$ \bm{H}, \bm{\Lambda} \gets (\bm{K}(\mathcal{X}),
		\bm{K}(\mathcal{Y})) $  \tcp*{eigen-decomposition~\cref{eq:KPCA}}
	}
	\caption{MV-RKM hyperparameter tuning with grid-search}
	\label{algo:dual_rrkm_train}
\end{algorithm}

\begin{algorithm}[ht]
	\DontPrintSemicolon
	\KwIn{Initial lagged vector: $\bm{x}_0 \in \mathbb{R}^{(p+1)d} $, time-steps $n'$}

	\KwOut{Recursive forecasts $\{\bm{y}_0, \ldots , \bm{y}_{n'}\}$}

	\While{$i \leq n'$}{
	\If{$i \neq 0$}{
		$\bm{x}_i \gets \{ \bm{y}_{i-1} , \bm{x}_{i-1}, \ldots, \bm{x}_{i-p} \}$ \;
	}
	Compute $\bm{k}_{\bm{x}}(\bm{x}_i) $ \tcp*{similarity vector $\mathcal{O}(n)$}
	Get $\bm{h}_i$ (see~\cref{eq:h_i_from_x}) \tcp*{latent vector $\mathcal{O}(1)$}
	\eIf{$\psi(\cdot)$ is known \& invertible}{
		$ \bm{y}_i \gets $ solve~\cref{eq:phi_y_hat} \tcp*{$\mathcal{O}(1)$}
	}{
		Compute $\bm{k}_y(\bm{y}_i)$ \tcp*{similarity vector $\mathcal{O}(n)$}
		$ \bm{y}_i \gets $ solve~\cref{eq:preimage_optimization} \tcp*{pre-image problem}
	}
	i++
	}
	\caption{MV-RKM recursive prediction}
	\label{algo:dual_rrkm_pred}
\end{algorithm}

\subsection{Computational complexity}

The eigendecomposition for the training (\cref{eq:KPCA}) requires slightly less than $\mathcal{O}(n^{3})$ operations, since the matrices are hermitian.
The complexity of the predictions in the latent space is
$\mathcal{O}(n)$ (\cref{eq:h_i_from_x}),
since it requires kernels $\bm{k}_{x}(\bm{x}_i)$.
Further, if the $k_y$ is a linear kernel, then the cost of prediction is constant i.e. $\mathcal{O}(1)$ (\cref{eq:phi_y_hat}).
Otherwise, one has to solve the pre-image problem and the complexity depends on the choice of pre-image method used. For instance with the kernel-smoother, it is $\mathcal{O}(n_r)$ and with kernel ridge regression, it is $\mathcal{O}(n^3)$, since it requires matrix inversion.

\section{Experiments}
In this section, we perform ablation studies on our proposed model on synthetic and real-world datasets to study the influence of different hyperparameters on the embeddings $\bm{H}$ and the predictions. Additionally, we show how visualizing the latent embeddings can help with explaining the data and model predictions.
Finally, the model is quantitatively compared against other standard time series models.

\subsection{Ablation study}

\begin{figure}[bh]
	\centering
	\centering
	\includegraphics[width=0.45\linewidth]{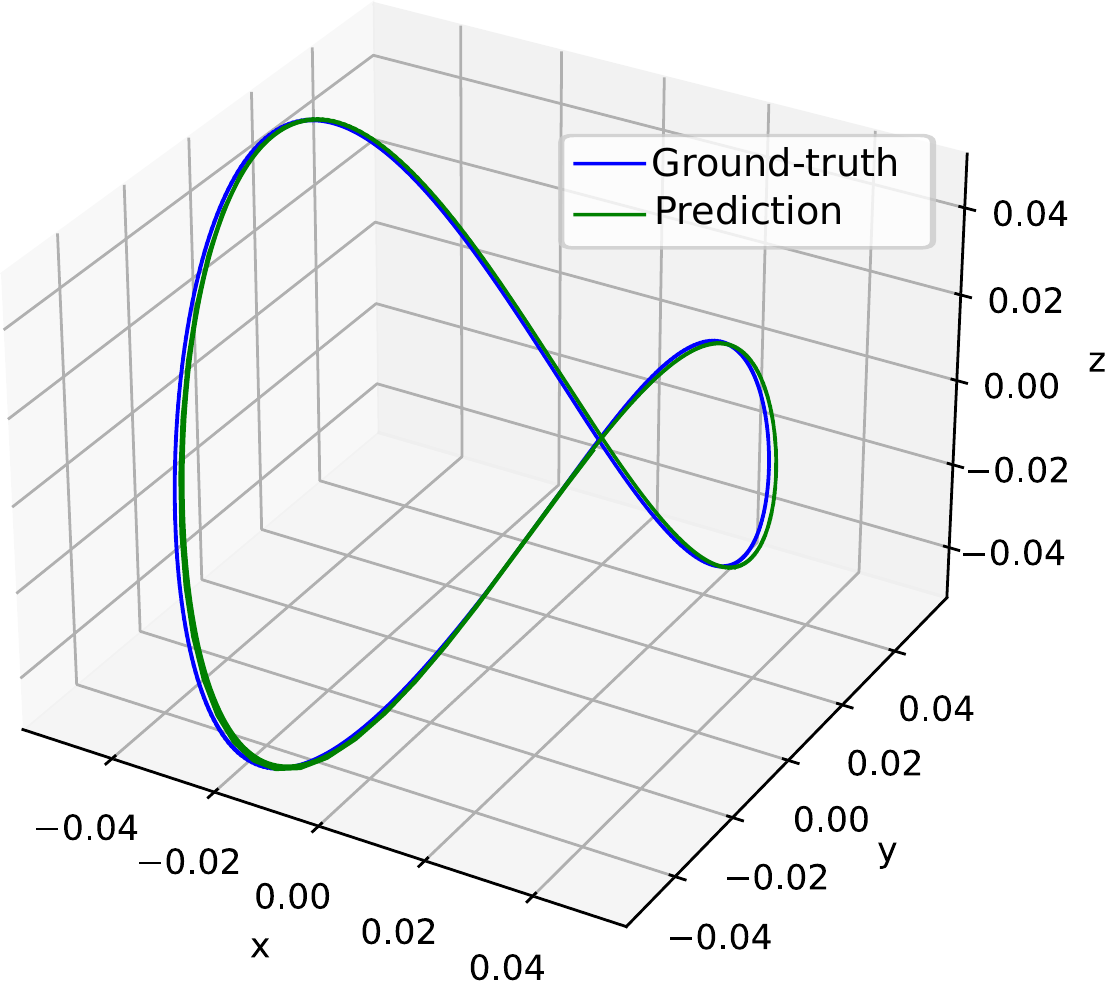}
	\includegraphics[width=0.45\linewidth]{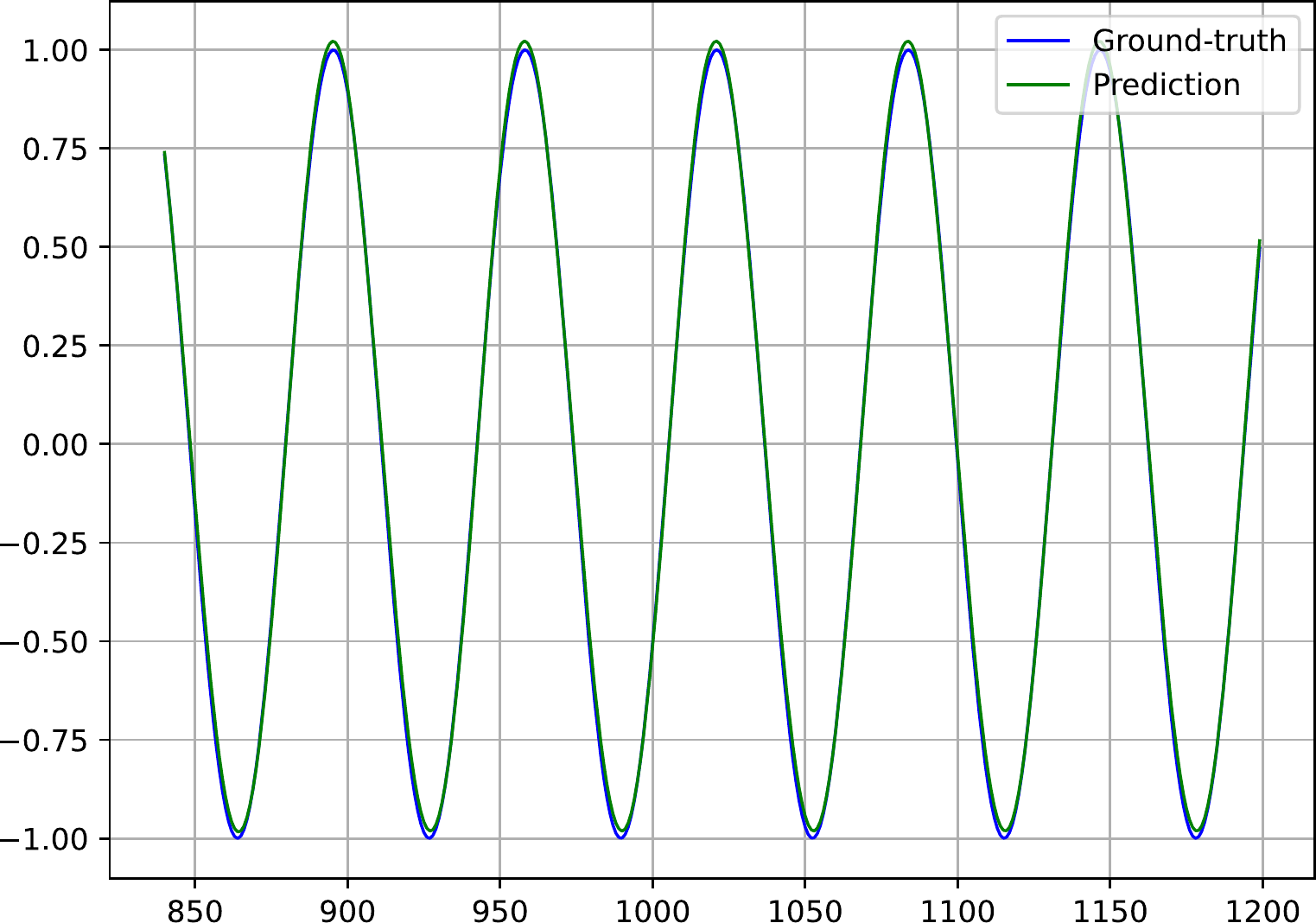}
	\caption{MV-RKM on $\sin$ wave. The embedded data points
		are \emph{temporally} de-correlated in the embedding space.}
	\label{fig:simple_sin}
\end{figure}

\textbf{Sine wave}. A simple sine wave is used to illustrate how the models learns the dynamics present in the input space, as latent representations.
After hyperparameter tuning, we obtain a closed-loop attractor in the latent space (ref.~\cref{fig:simple_sin}).
The forecasted embeddings of the unseen test set, indicated by the green curve, follow the same manifold as the embeddings of training set as indicated with blue curve.
When the forecasted embeddings are transformed back to input space, it becomes apparent that the model has learned the data correctly, reflected by the high accuracy of the forecasted data points.

\begin{figure}[h!]
	\centering
	\setlength{\tabcolsep}{0pt}
	\resizebox{\textwidth}{!}{
		\begin{tabular}{r c c}
			                                                         & {\footnotesize Predictions}
			                                                         & {\footnotesize Latent components}                                                                                                               \\
			\rotatebox[origin=c]{90}{\footnotesize Sine (lag 83)}    & \tabfigure{width=6.5cm, height=4cm}{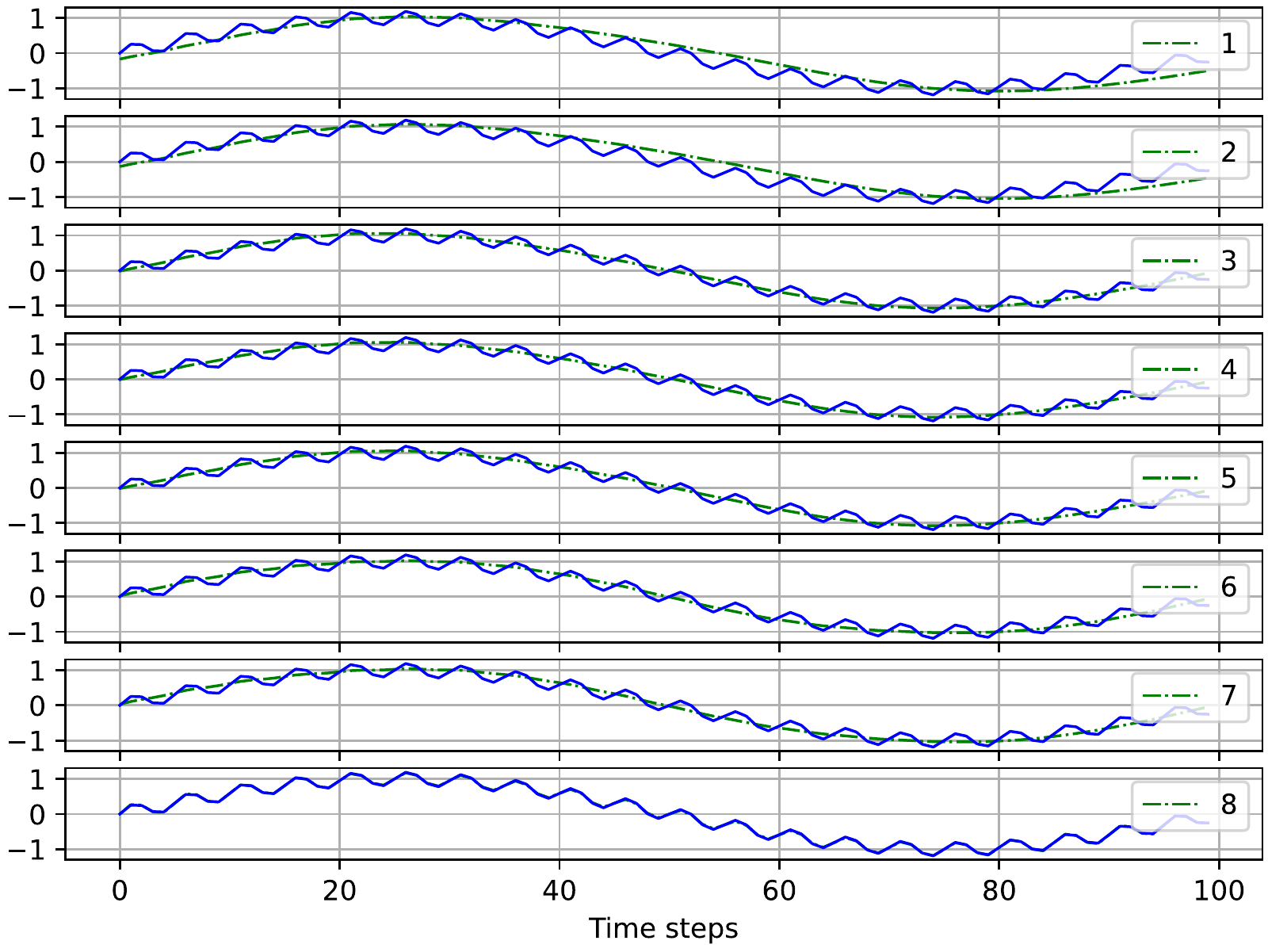}       & \tabfigure{width=6.5cm, height=4cm}{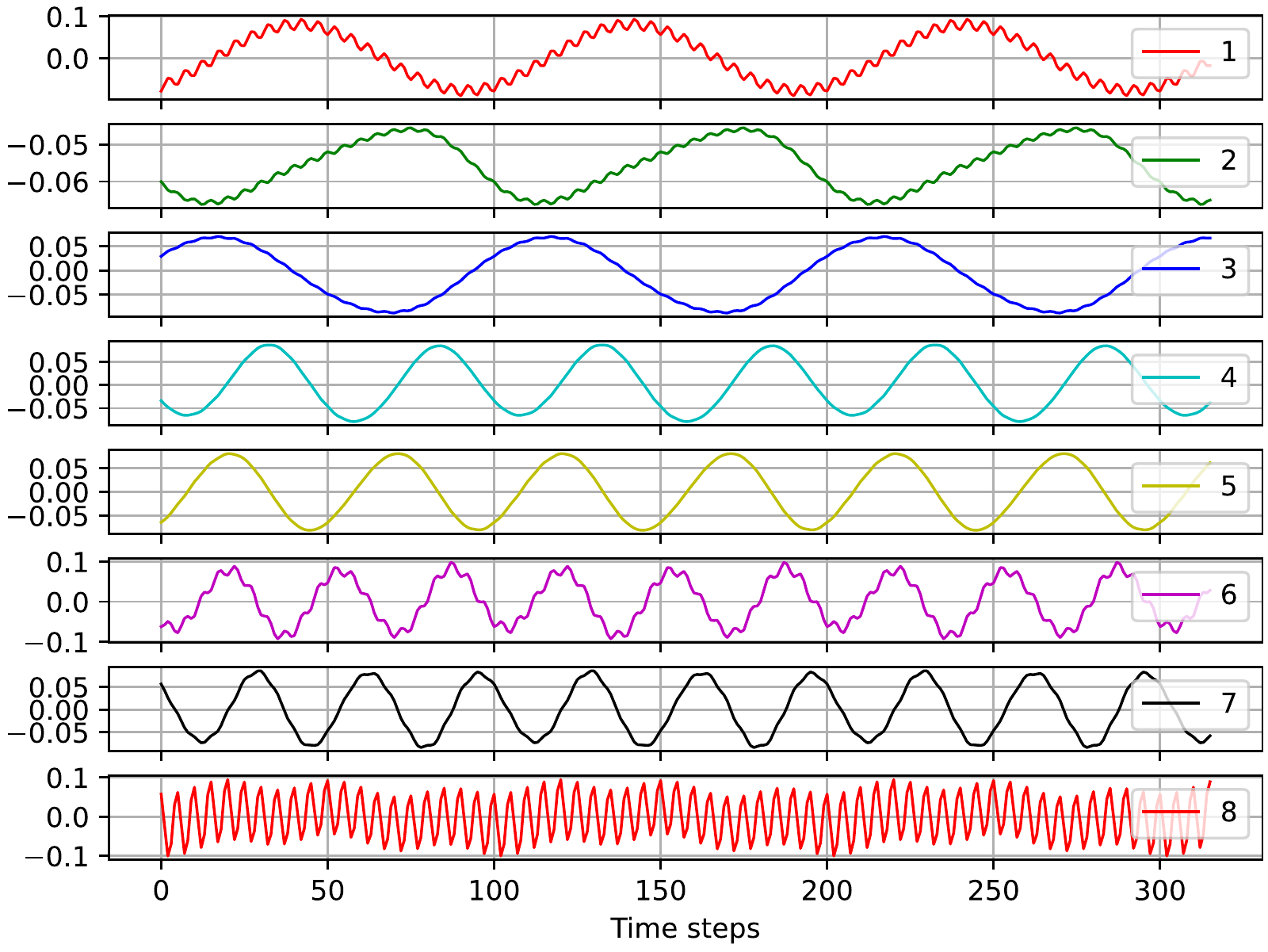}       \\
			\rotatebox[origin=c]{90}{\footnotesize Sine (lag 10)}    & \tabfigure{width=6.5cm, height=4cm}{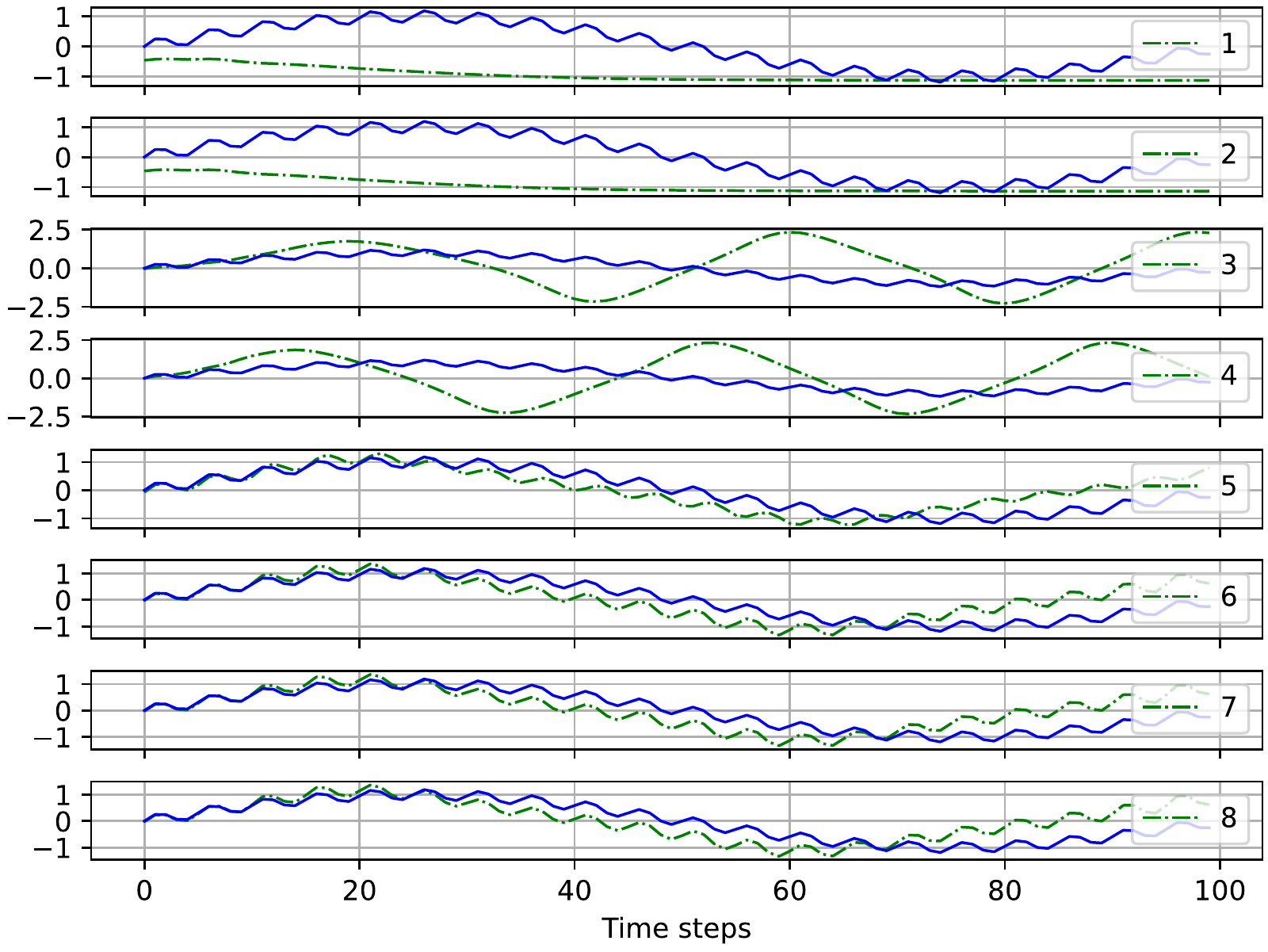} & \tabfigure{width=6.5cm, height=4cm}{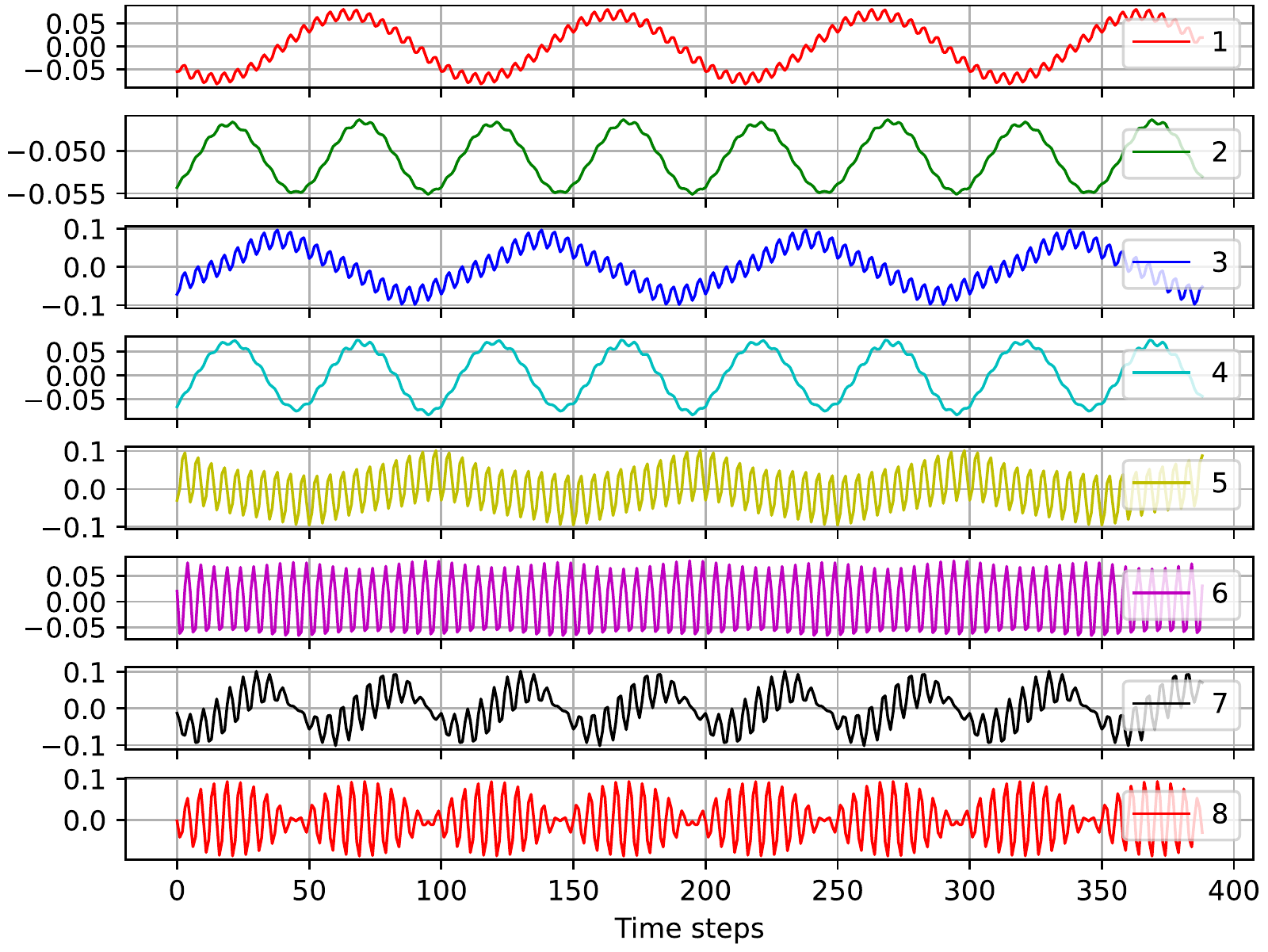} \\
			\rotatebox[origin=c]{90}{\footnotesize SantaFe (lag 70)} & \tabfigure{width=6.5cm, height=3.5cm}{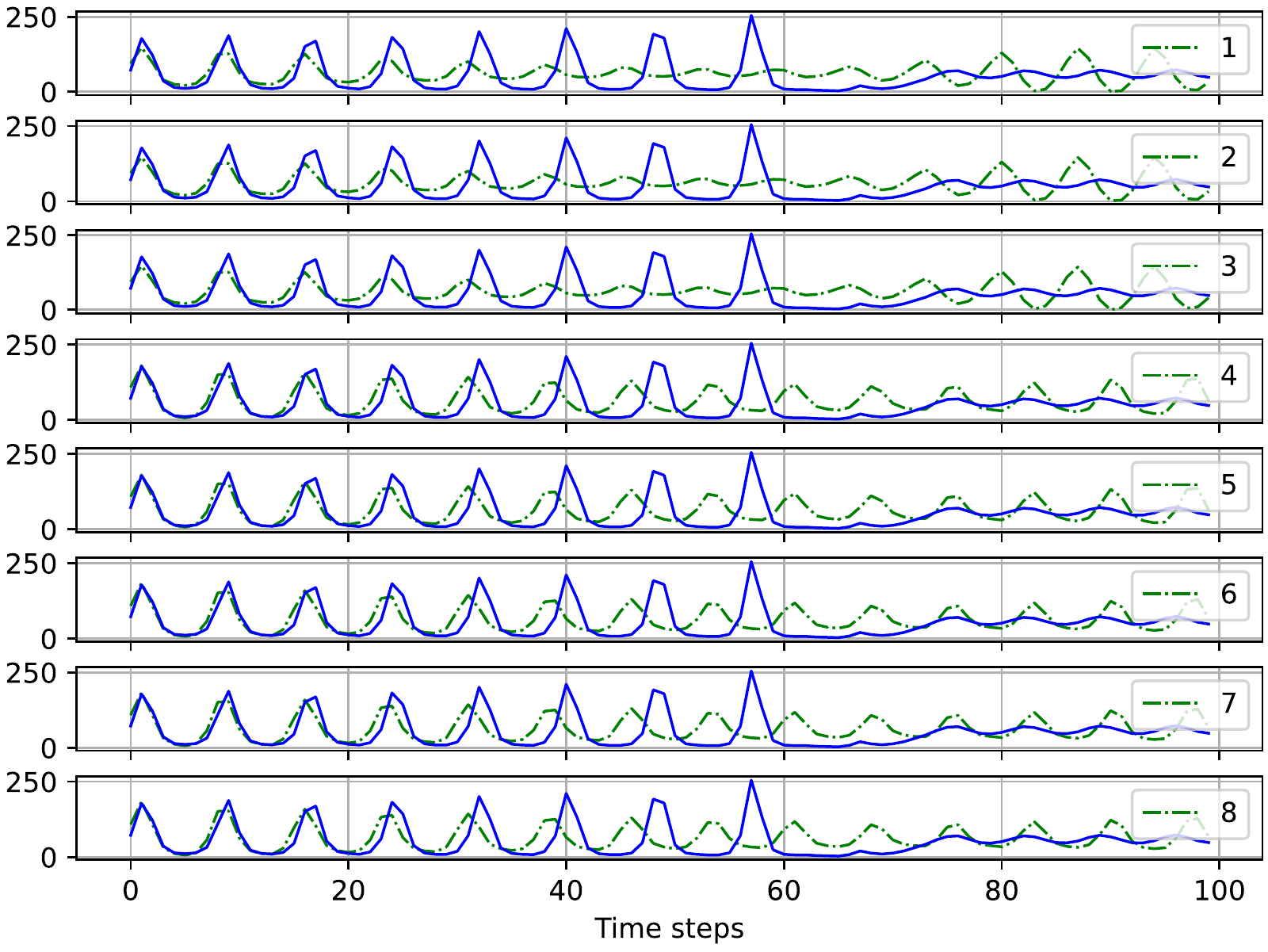}  & \tabfigure{width=6.5cm, height=3.5cm}{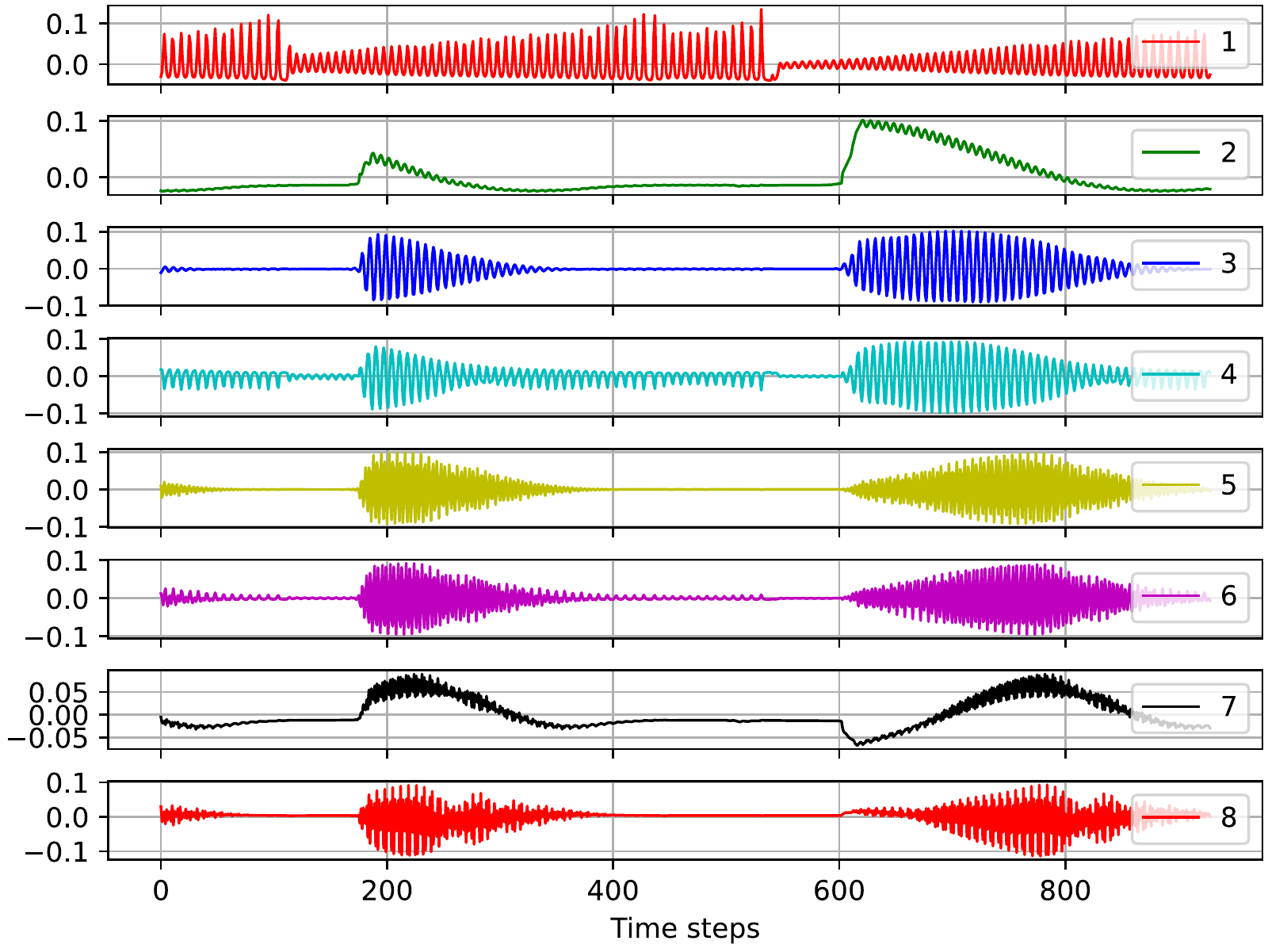}
		\end{tabular}
	}
	\caption{Predictions (left) and components (right) for the
    		sum of sine and santafe experiments.
    		More components correspond to a better prediction. A lower lag for the same parameters
            generally generate higher frequency components and more components are required for the same
            performance.
            }
	\label{fig:preds_and_latent_components}
\end{figure}

\begin{figure}
	\centering %
	\begin{subfigure}[b]{0.45\textwidth}
		\includegraphics[width=\linewidth]{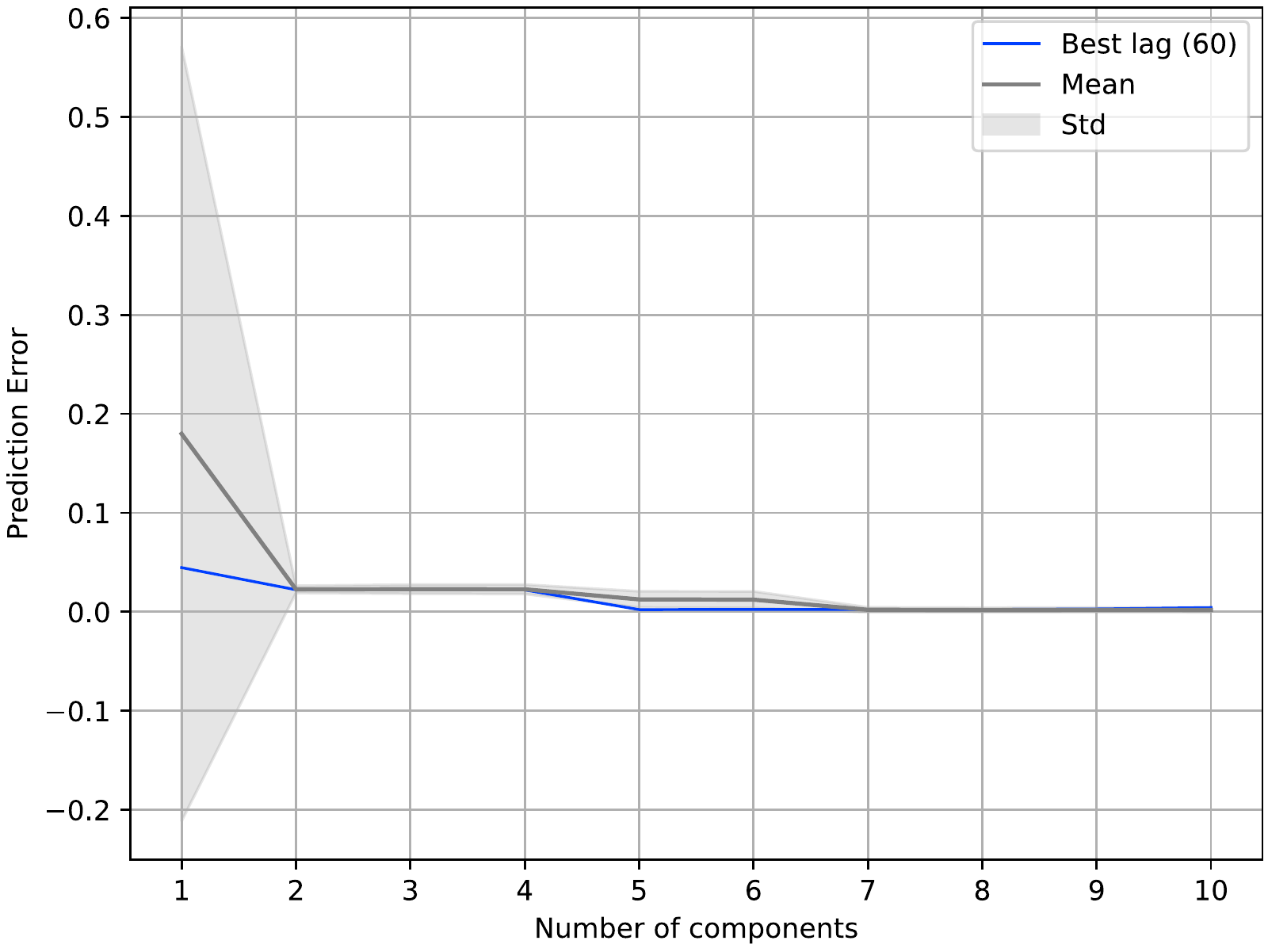}
		\caption{Sum of sine}
		\label{fig:latent_components}
	\end{subfigure}
	\begin{subfigure}[b]{0.45\textwidth}
		\includegraphics[width=\linewidth]{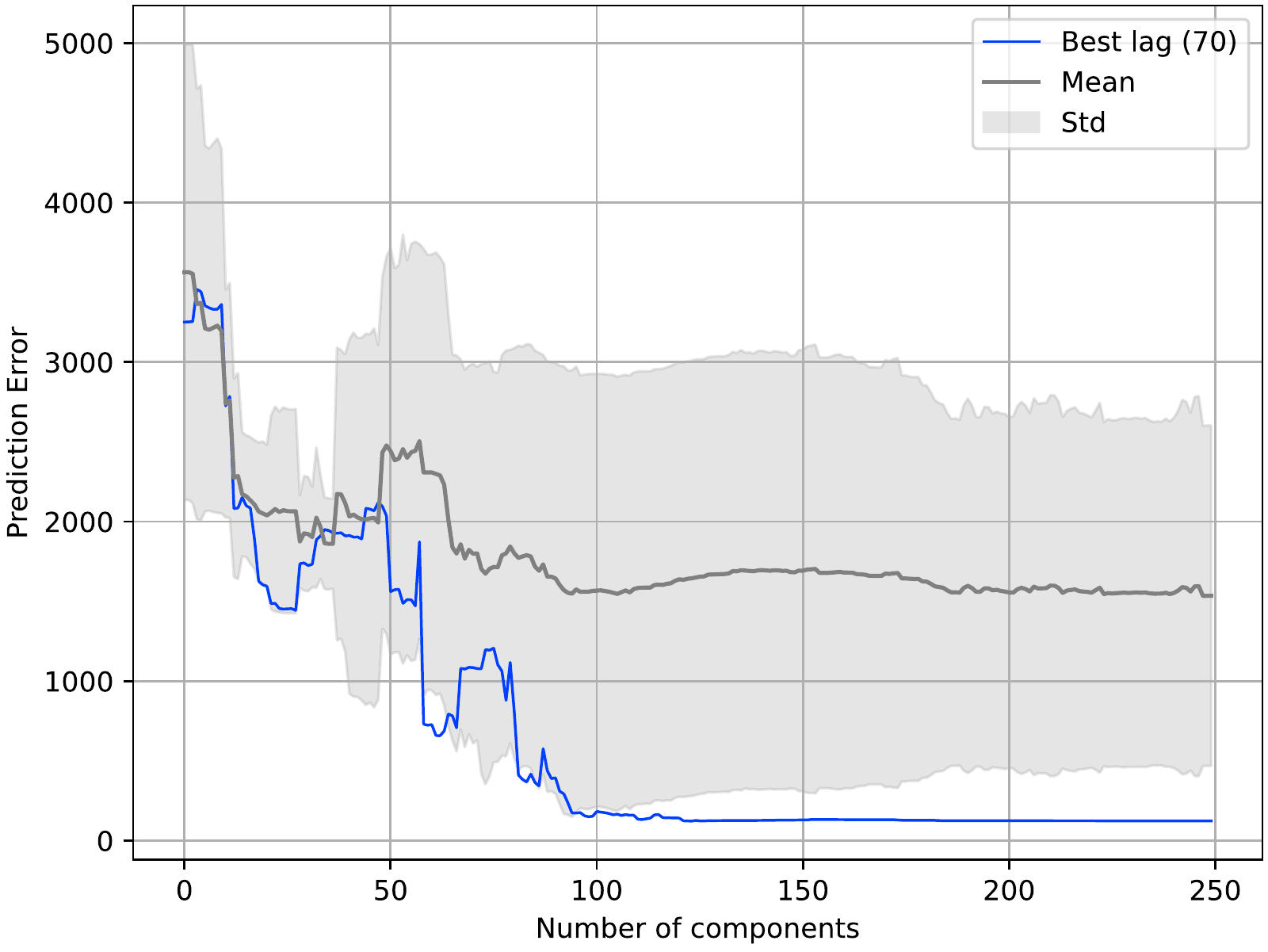}
		\caption{SantaFe}
		\label{fig:latent_components_mean_std}
	\end{subfigure}
	\caption{Prediction error with different number of components on the sum of sines (left) and SantaFe (right) datasets. The blue curve indicates the prediction of the model for the best lag. The gray curve indicates the mean prediction error over multiple lag values with corresponding standard deviation. It can be seen that more components indicate a better reconstruction until a plateau is reached.}
	\label{}
\end{figure}

\textbf{Sum of sines}. A synthetic dataset was created as the sum of two since waves as follows:
$\sum_{i=1}^{n} \alpha_i \sin(2
	\pi \omega_i t + a_i)$, where $ n=2$, $ \alpha_i \in \{ 1, 0.2 \}$, $\omega_i \in
	\{ 1, 20 \}$, $a_i \in \{0, 0 \}  $.
Using this dataset, we illustrate how the predictions of the model can be explained using the orthogonal latent components.
On~\cref{fig:preds_and_latent_components}, the latent components are visualised on the right.
On the left, the predictions of the model are shown where iteratively more and more of the plotted components are used.
Visually, it can be seen that more components correspond with a more accurate forecast.
This is further shown by plotting the error, in terms of the mean squared error, as done on~\cref{fig:latent_components}.
When visualised, the orthogonal components offer a measure of explainability and show how the model decomposes the signal into uncorrelated components.
Generally, we observe that higher frequency components are extracted for lower
lag values and lower frequency components for high lag values (see first and
second row in~\cref{fig:preds_and_latent_components}).
Empirically, it was found that a lower lag means that a higher number of components are required to fit the model sufficiently well.

\textbf{SantaFe}.
The same experiment was performed on the SantaFe data, which is a real world
chaotic uni-variate time series dataset.
We can observe the same effects as with the sine experiments
in~\cref{fig:preds_and_latent_components} (last row).
The predictions improve given increasingly more of the visualised components. Further, the components can be observed to be able to capture the jumps in the chaotic time series, given a certain lag.
It can be observed that, given a sufficiently large lag, more components correlate with a lower prediction error (see~\cref{fig:latent_components_mean_std}). This effect is similar as in principal component analysis where only the top components are needed for a good reconstruction of the data.

\subsection{Real-world experiments}
\textbf{Baselines}: To provide a benchmark study, we consider six baseline models: Recurrent Neural Networks (RNN),
Long-Short Term Memory networks (LSTM), Autoregressive Moving Average (ARMA), Kernel-Recursive Least-Square~\citep{krls_tracker} (KRLS), non-linear autoregressive formulation of the Least-Squares Support Vector Machines~\citep{suykens_least_2002} (LS-SVM), Recurrent Restricted Kernel Machine~\citep{rrkm_esann} (RRKM).
Further we consider two variants of the proposed model i.e. when $k_y(\bm{y}_i, \bm{y}_j) = \langle \bm{y}_i , \bm{y}_j \rangle $ and $k_y(\bm{y}_i, \bm{y}_j) = \exp (- \lVert \bm{y}_i - \bm{y}_j \rVert^2 / {2 \sigma_y^2} ) $; with $k_x(\bm{x}_i, \bm{x}_j) = \exp (- \lVert \bm{x}_i - \bm{x}_j \rVert^2 / {2 \sigma_x^2} ) $ in both cases.

\begin{table}
	\caption{Mean squared error on the forecasted
		data. Standard deviation for 10 iterations between brackets for the
		stochastic models.}
	\label{tab:mse_1_step}
	\centering
	\resizebox{\textwidth}{!}{%
		\begin{tabular}{@{}lllllllll@{}}
			\multicolumn{1}{l}{\multirow{2}{*}{\textbf{Data}}}   &
			\multicolumn{1}{c}{\multirow{2}{*}{\textbf{RNN}}}    &
			\multicolumn{1}{c}{\multirow{2}{*}{\textbf{ARMA}}}   &
			\multicolumn{1}{c}{\multirow{2}{*}{\textbf{KRLS}}}   &
			\multicolumn{1}{c}{\multirow{2}{*}{\textbf{LSTM}}}   &
			\multicolumn{1}{c}{\multirow{2}{*}{\textbf{LS-SVM}}} &
			\multicolumn{1}{c}{\multirow{2}{*}{\textbf{RRKM}}}   &
			\multicolumn{2}{c}{\textbf{RKM-MV}}                    \\ \cmidrule(lr){8-9}
			                                                     &
			                                                     &
			                                                     &
			                                                     &
			                                                     &
			                                                     &
			                                                     &
			\multicolumn{1}{c}{($k_y$ rbf)}                      &
			\multicolumn{1}{c}{($k_y$ lin)}                        \\ \midrule
			SantaFe                                              &
			$3037.17~(\pm625.36)$                                &
			$2224.55$                                            &
			$416.17$                                             &
			$2272.13~(\pm328.87)$                                &
			$113.78$                                             &
			{$119.06$}                                           &
			$\bm{90.23}$                                         &
			127.83                                                 \\
			Lorenz                                               &
			$386.01~(\pm51.53)$                                  &
			$412.28$                                             &
			$273.00$                                             &
			$331.49~(\pm61.30)$                                  &
			$\bm{131.87}$                                        &
			$247.19$                                             &
			$182.65$                                             &
			163.659                                                \\
			Chickenpox                                           &
			$34329.95~(\pm9513.07)$                              &
			$23571.35$                                           &
			$19505.99$                                           &
			$22296.24~(\pm3097.18)$                              &
			$15541.28$                                           &
			$20716.91$                                           &
			$\bm{15479.93}$                                      &
			16807.56                                               \\
			Energy                                               &
			$16002.11~(\pm 1809.89)$                             &
			$24797.40 $                                          &
			$11994.98$                                           &
			$14781.65~(\pm 3126.80)$                             &
			$10635.00$                                           &
			$12764.097$                                          &
			$\bm{10385.07}$                                      &
			10474.06                                               \\
			Turbine                                              &
			$2401.12~(\pm 644.53)$                               &
			$1317.67$                                            &
			$1193.82$                                            &
			$1817.38~(\pm 552.21)$                               &
			$1226.87$                                            &
			$1299.915$                                           &
			$\bm{1126.55}$                                       &
			1291.72
		\end{tabular}%
	}
\end{table}

\textbf{Datasets}: We consider both uni-variate and multi-variate time series: two chaotic time series (SantaFe laser\footnote{\url{https://rdrr.io/cran/TSPred/man/SantaFe.A.html}}~\citep{santafe},
Lorenz attractor\footnote{\url{https://www.openml.org/search?type=data&sort=runs&id=42182&status=active}})
and three real world datasets
(Chickenpox\footnote{\url{https://archive.ics.uci.edu/ml/datasets/Hungarian+Chickenpox+Cases}}~\citep{chickenpox},
Belgian appliances energy consumption\footnote{\url{https://archive.ics.uci.edu/ml/datasets/Appliances+energy+prediction}}~\citep{appliances_energy}
and
Gas turbine\footnote{\url{https://archive.ics.uci.edu/ml/datasets/Gas+Turbine+CO+and+NOx+Emission+Data+Set}}~\citep{gas_turbine}). See~\ref{datasets_and_hyperparams} for further details on these datasets.

On each dataset and method, hyperparameter tuning
has been performed with grid-search and the result of the best set of parameters, quantified
as the mean squared error, is shown in~\cref{tab:mse_1_step}.
For all methods, the entire  validation set is forecasted, in \emph{recursive} fashion, with the
initial vector as the last point of the training set.
The scores for RNN, ARMA and RRKM are taken from~\cite{rrkm_esann} since the experiment setting remains same.
From the table, we see that the proposed model RKM-MV outperforms the baselines on most datasets except Lorenz where the LS-SVM has lower error than RKM-MV.
Also note that these two models are closest in performance, compared to the rest, since both are in NAR setting.
However the solution to LS-SVM requires solving a linear system (ref. \ref{sec:ls-svm}) in contrast with the eigen-decomposition of the proposed model. Another important difference is in the prediction schemes.

Moreover, LS-SVM outperforms RKM-MV when $k_y$ is linear on most datasets except the Belgian appliances energy dataset. 
However, the ability to include non-linearity on $k_y$ is an added advantage of the RKM-MV model which allows it to outperform other methods.

\section{Conclusion}
In this work, we introduced a time series forecasting model based on multi-view kernel PCA.
This framework provides new insights into time series modeling such as latent space dynamics and novel relations between kernel PCA and time series forecasting.
For future work, we believe a
further extension of this model can be made to a multi-source setting. For example,
when the data is coming from multiple weather stations and a forecast has to made for all stations simultaneously.
Custom kernels can be used to better capture the relationship between data-points.
Moreover, existing techniques in kernel methods literature can be leveraged
to improve the computational complexity such as the Nystr\"om approximation
or random Fourier features.
Besides the topics mentioned in this work, the framework can be further explored towards
other tasks involving time series such as denoising, handling missing values and
clustering.

\subsection*{Acknowledgements}
European Research Council under the European
Union's Horizon 2020 research and innovation programme: ERC Advanced Grants
agreements E-DUALITY(No 787960) and Back to the Roots (No 885682). This paper
reflects only the authors' views and the Union is not liable for any use that
may be made of the contained information. Research Council KUL: Optimization
frameworks for deep kernel machines C14/18/068; Research Fund (projects
C16/15/059, C3/19/053, C24/18/022, C3/20/117, C3I-21-00316); Industrial Research
Fund (Fellowships 13-0260, IOFm/16/004, IOFm/20/002) and several Leuven Research
and Development bilateral industrial projects. Flemish Government Agencies: FWO:
projects: GOA4917N (Deep Restricted Kernel Machines: Methods and Foundations),
PhD/Postdoc grant, EOS Project no G0F6718N (SeLMA), SBO project S005319N,
Infrastructure project I013218N, TBM Project T001919N, PhD Grant (SB/1SA1319N);
EWI: the Flanders AI Research Program; VLAIO: CSBO (HBC.2021.0076) Baekeland PhD
(HBC.20192204). This research received funding from the Flemish Government (AI
Research Program). Other funding: Foundation `Kom op tegen Kanker', CM
(Christelijke Mutualiteit).

\appendix

\section{Note on kernel centering}
\paragraph{Centering Kernel Matrix} Let
$\Phi = [ \phi(\bm{x}_1), \ldots,  \phi(\bm{x}_n)] $ and the sample mean be
$\bm{\mu}_{\phi} = \frac{1}{n} \sum_{i=1}^{n} \phi(\bm{x}_i) = \frac{1}{n} \Phi
	\mathbbm{1}_{n}$, where $\mathbbm{1}_{n} = [1, \ldots , 1]^\top$. Let
$\tilde{\phi}(\bm{x}_i) = \phi(\bm{x}_i) - \bm{\mu}_{\phi}$ denote the centered
feature vector. Then
\begin{empheq}{align}
	\tilde{k}(\bm{x}_i , \bm{x}_i)  = & k(\bm{x}_i , \bm{x}_i) -
	\langle \phi(\bm{x}_i), \bm{\mu}_{\phi} \rangle_{\mathscr{H}}- \langle
	\bm{\mu}_{\phi},  \phi(\bm{x}_i) \rangle_{\mathscr{H}}- \langle
	\bm{\mu}_{\phi},  \bm{\mu}_{\phi} \rangle_{\mathscr{H}} \nonumber
	\\
	\tilde{k}(\bm{x}_i , \bm{x}_i)  = & k(\bm{x}_i , \bm{x}_i) - \langle
	\phi(\bm{x}_i),  \Phi \mathbbm{1}_{n} \rangle_{\mathscr{H}}\frac{1}{n}
	-\frac{1}{n} \langle   \Phi \mathbbm{1}_{n},  \phi(\bm{x}_i)
	\rangle_{\mathscr{H}}+ \frac{1}{n}\langle  \Phi \mathbbm{1}_{n},  \Phi
	\mathbbm{1}_{n} \rangle_{\mathscr{H}} \frac{1}{n} \nonumber
\end{empheq}
Composing the above in the matrix form $\forall i$, gives the following
\begin{empheq}{align}
	\bm{\tilde{K}} = & \bm{K} - \bm{K} \mathbbm{1}
	\mathbbm{1}^{\top} \frac{1}{n}- \frac{1}{n}\mathbbm{1}  \mathbbm{1}^{\top}
	\bm{K}  +  \frac{1}{n^2}\mathbbm{1}  \mathbbm{1}^{\top} \bm{K}\mathbbm{1}
	\mathbbm{1}^{\top} \nonumber                                         \\
	\bm{\tilde{K}} = & \big(\mathbb{I} - \mathbbm{1}  \mathbbm{1}^{\top}
	\frac{1}{n} \big) \bm{K}  \big(\mathbb{I} - \mathbbm{1}  \mathbbm{1}^{\top}
	\frac{1}{n} \big) \label{eq:kernel_centering}
\end{empheq}
\paragraph{Centering Kernel Vector} Using the same sample statistic
$\bm{\mu}_{\phi}$, a kernel vector representing the similarities
$\bm{k}_{\bm{x}_i}(\bm{x}^\star) = [{k}(\bm{x}_1, \bm{x}^\star), \ldots ,
	{k}(\bm{x}_n, \bm{x}^\star)]^\top$   can be centered as follows : \begin{align}
	\bm{\tilde{k}}_{\bm{x}_i}(\bm{x}^\star) = \bm{k}_{\bm{x}_i}(\bm{x}^\star) -
	\bm{K} \mathbbm{1}  \frac{1}{n}- \frac{1}{n}\mathbbm{1}  \mathbbm{1}^{\top}
	\bm{k}_{\bm{x}_i}(\bm{x}^\star)  +  \frac{1}{n^2}\mathbbm{1}
	\mathbbm{1}^{\top} \bm{K}\mathbbm{1} \label{eq:kernel_vector_centering}
\end{align}

\section{Datasets and Hyperparameters \label{datasets_and_hyperparams}}

We refer to~\cref{Table:dataset}  for more  details  on  model  architectures, datasets and hyperparameters used in this paper. The PyTorch library (double precision) in Python was used. See Algorithm~\ref{algo:dual_rrkm_train} for training the MV-RKM model.

\begin{table}[ht]
	\caption{Datasets used for the experiments. $N$  is the number of training samples, $d$ the input dimension, $s$ the subspace dimension and $m$ the minibatch size.}
	\label{Table:dataset}
	\centering
	\begin{tabular}{lllll}
		\toprule
		\textbf{Dataset} & $N_{\text{train}}$ & $N_{\text{test}}$ & $d$ \\ \midrule
		SantaFe          & 1000               & 100               & 1   \\
		Lorenz           & 2801               & 1200              & 3   \\
		Chickenpox       & 418                & 104               & 20  \\
		Energy           & 4000               & 1000              & 28  \\
		Gas Turbine      & 5929               & 1482              & 11  \\ \bottomrule
	\end{tabular}
\end{table}

Lorenz attractor is defined by the following dynamical system:
\begin{empheq}[left=\empheqlbrace]{align}
	\dot{x}(t) &= - a x + a y \nonumber \\
	\dot{y}(t) &= - xz + rx - y \nonumber \\
	\dot{z}(t) &= xy - bz. \nonumber
\end{empheq}
For our simulations, we set $a=10$, $r=28$ and $b=2.667$ with initial condition $x(0)=1.0$, $y(0) =-1.0$ and $z(0)=1.05$ and the above equations are solved using first-order finite difference scheme.

\begin{figure}
	\centering
	\begin{subfigure}[b]{0.45\textwidth}
		\centering
		\def\svgwidth{\linewidth}
		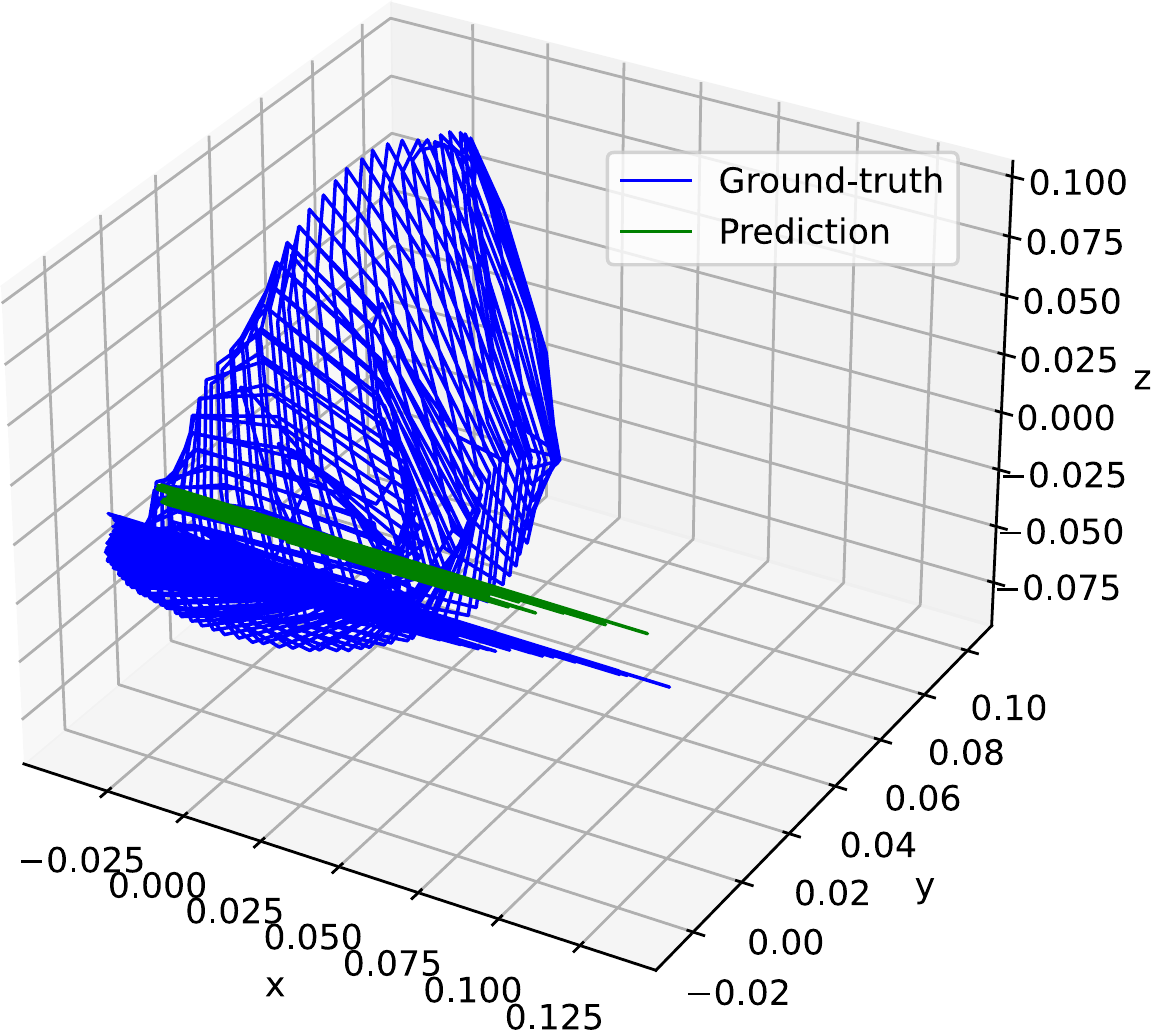
		\caption{Latent space}
		\label{fig:santafe_latent_distribution}
	\end{subfigure} 
	\begin{subfigure}[b]{0.45\textwidth}
		\centering
		\def\svgwidth{\linewidth}
		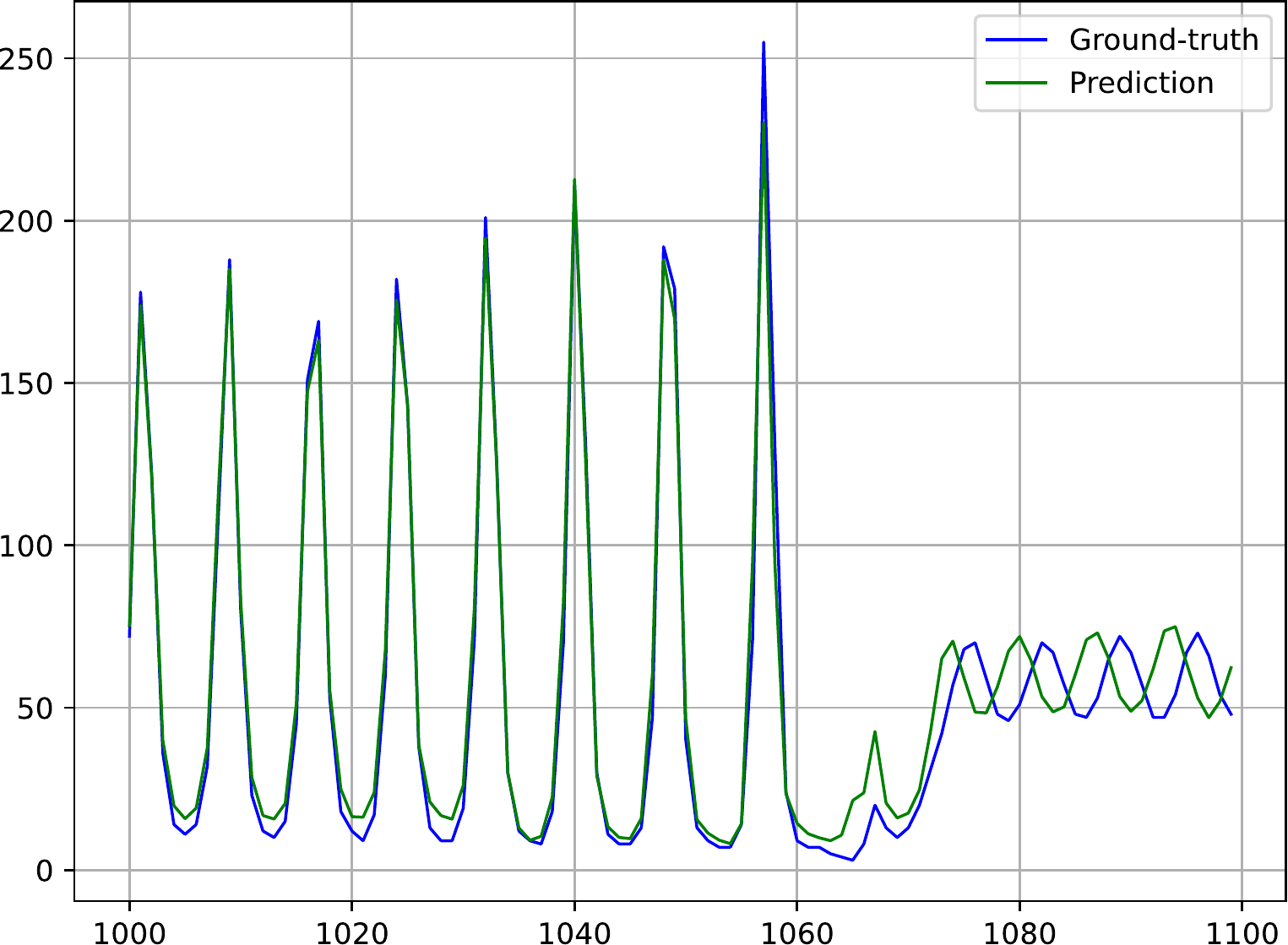
		\caption{Predictions}
		\label{fig:santafe_prediction}
	\end{subfigure}
	\begin{subfigure}[b]{0.45\textwidth}
		\centering
		\def\svgwidth{\linewidth}
		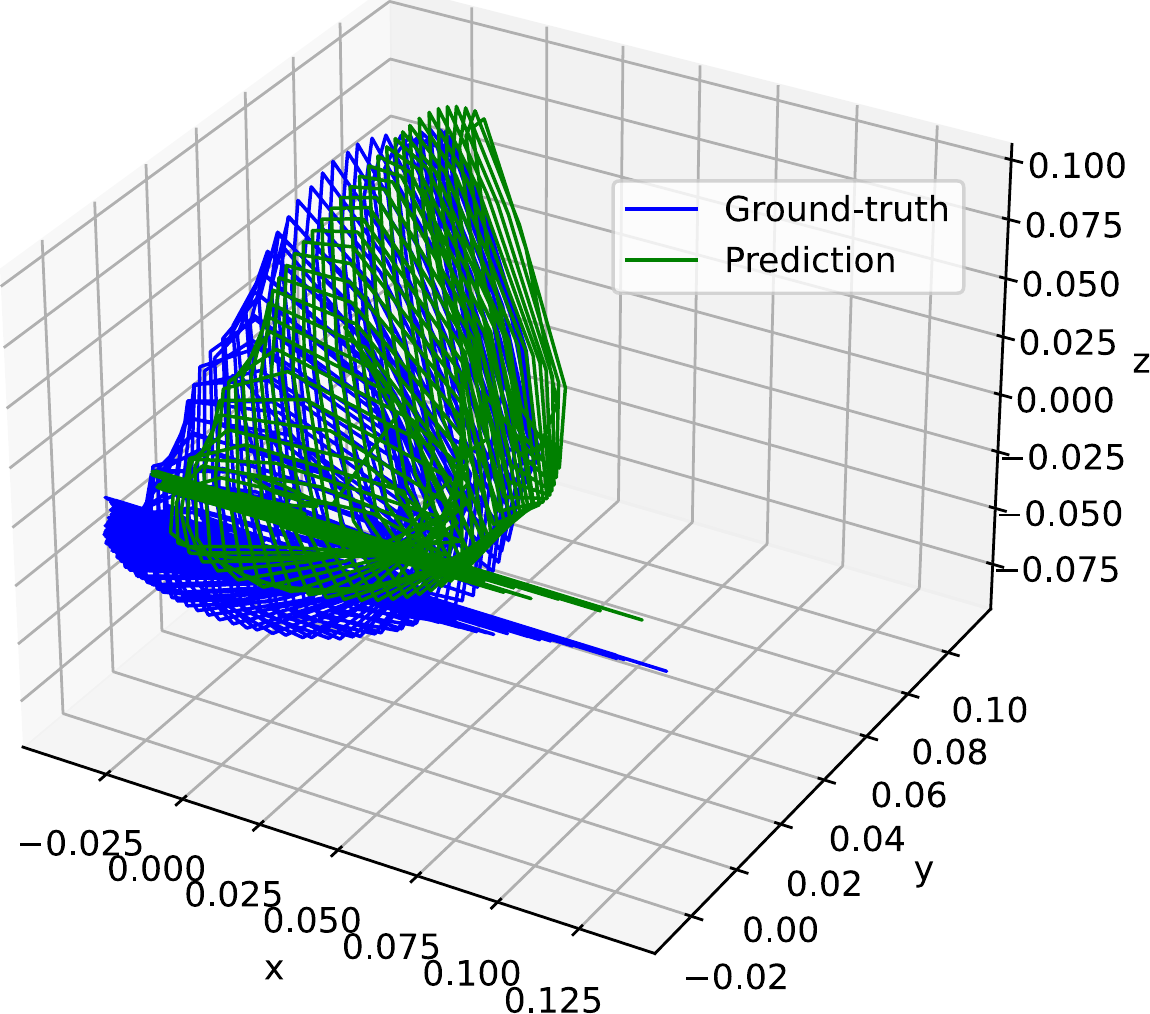
		\caption{Latent space}
		\label{fig:santafe_latent_distribution_long_term}
	\end{subfigure} 
	\begin{subfigure}[b]{0.45\textwidth}
		\centering
		\def\svgwidth{\linewidth}
		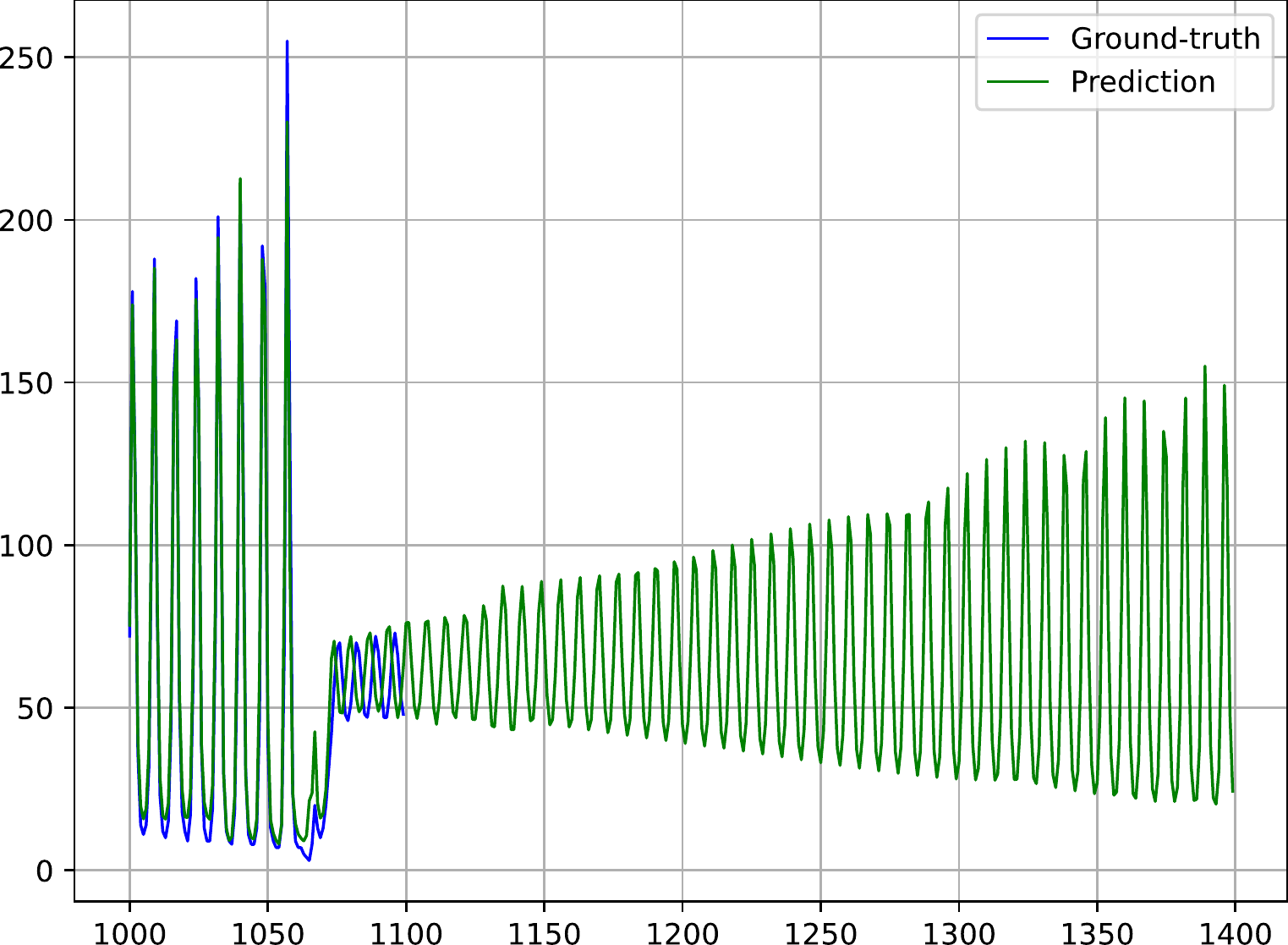
		\caption{Predictions}
		\label{fig:santafe_prediction_long_term}
	\end{subfigure}
 
	\caption{MV-RKM (rbf $k_x (\sigma=2.1856)$, linear $k_y$, $s=144$) on the SantaFe chaotic laser dataset showing short-term and long-term predictions in the first and second row respectively.}
	\label{fig:santafe_latent_distribution_and_prediction}
\end{figure}

\section{Least-Squares Support Vector Machine}
\label{sec:ls-svm}
We consider the non-linear auto-regressive setting for LS-SVM regression model: $ \bm{y} = \sum_{i=1}^{n}\bm{\alpha}_i k_x ( \bm{x}_i, \bm{x} ) + \bm{b} $, where
the parameters $\bm{\alpha}, \bm{\beta}$ are found by solving the following linear system:
$ \left[ \begin{array}{c|c}
			0          & \bm{1}_{n}^\top                              \\ \hline
			\bm{1}_{n} & \bm{K}(\mathcal{X}) + \gamma^{-1} \mathbb{I}
		\end{array} \right]
	\left[ \begin{array}{c}
			\bm{b} \\ \hline
			\bm{\alpha}
		\end{array} \right]
	=
	\left[ \begin{array}{c}
			0 \\ \hline
			\bm{y}
		\end{array}
		\right]$.
For details on the derivation from a constrained optimization problem, we refer to~\cite[p. 98]{suykens_least_2002}.

\FloatBarrier

\bibliographystyle{unsrt}

\begin{thebibliography}{10}

\bibitem{vapnik95}
Vladimir~N. Vapnik.
\newblock {\em The nature of statistical learning theory}.
\newblock Springer-Verlag New York, Inc., 1995.

\bibitem{Scholkopf2001}
Bernhard Scholkopf and Alexander~J. Smola.
\newblock {\em Learning with Kernels: Support Vector Machines, Regularization,
  Optimization, and Beyond}.
\newblock MIT Press, Cambridge, MA, USA, 2001.

\bibitem{suykens_least_2002}
Johan A.~K. Suykens, Tony Van~Gestel, Jos De~Brabanter, Bart De~Moor, and Joos
  Vandewalle.
\newblock {\em Least Squares Support Vector Machines}.
\newblock World Scientific, River Edge, NJ, January 2002.

\bibitem{gp_machine_learning_Rasmussen06}
Carl~Edward Rasmussen and Christopher K.~I. Williams.
\newblock {\em Gaussian processes for machine learning.}
\newblock Adaptive computation and machine learning. MIT Press, 2006.

\bibitem{kpca}
Bernhard Schölkopf, Alexander Smola, and Klaus-Robert Müller.
\newblock Nonlinear component analysis as a kernel eigenvalue problem.
\newblock {\em Neural Computation}, 10(5):1299--1319, 1998.

\bibitem{kfda}
S.~Mika, G.~Ratsch, J.~Weston, B.~Scholkopf, and K.R. Mullers.
\newblock Fisher discriminant analysis with kernels.
\newblock In {\em Neural Networks for Signal Processing IX: Proceedings of the
  1999 IEEE Signal Processing Society Workshop (Cat. No.98TH8468)}, pages
  41--48, 1999.

\bibitem{Oreshkin2020N-BEATS}
Boris~N. Oreshkin, Dmitri Carpov, Nicolas Chapados, and Yoshua Bengio.
\newblock N-beats: Neural basis expansion analysis for interpretable time
  series forecasting.
\newblock In {\em International Conference on Learning Representations}, 2020.

\bibitem{Temporal_Fusion_Transformers_2021}
{Bryan Lim and Sercan Ö. Arik and Nicolas Loeff and Tomas Pfister}.
\newblock Temporal fusion transformers for interpretable multi-horizon time
  series forecasting.
\newblock {\em International Journal of Forecasting}, 37(4):1748--1764, 2021.

\bibitem{fedformer_zhou22}
Tian Zhou, Ziqing Ma, Qingsong Wen, Xue Wang, Liang Sun, and Rong Jin.
\newblock {FED}former: Frequency enhanced decomposed transformer for long-term
  series forecasting.
\newblock In Kamalika Chaudhuri, Stefanie Jegelka, Le~Song, Csaba Szepesvari,
  Gang Niu, and Sivan Sabato, editors, {\em Proceedings of the 39th
  International Conference on Machine Learning}, volume 162 of {\em Proceedings
  of Machine Learning Research}, pages 27268--27286. PMLR, 17--23 Jul 2022.

\bibitem{nhits_2022_arxiv}
Cristian Challu, Kin~G. Olivares, Boris~N. Oreshkin, Federico Garza, Max
  Mergenthaler-Canseco, and Artur Dubrawski.
\newblock N-hits: Neural hierarchical interpolation for time series
  forecasting, 2022.

\bibitem{Kernel_Methods_for_Deep_Learning_2009}
Youngmin Cho and Lawrence Saul.
\newblock Kernel methods for deep learning.
\newblock In Y.~Bengio, D.~Schuurmans, J.~Lafferty, C.~Williams, and
  A.~Culotta, editors, {\em Advances in Neural Information Processing Systems},
  volume~22. Curran Associates, Inc., 2009.

\bibitem{Deep_Kernel_Learning_wilson16}
Andrew~Gordon Wilson, Zhiting Hu, Ruslan Salakhutdinov, and Eric~P. Xing.
\newblock Deep kernel learning.
\newblock In Arthur Gretton and Christian~C. Robert, editors, {\em Proceedings
  of the 19th International Conference on Artificial Intelligence and
  Statistics}, volume~51 of {\em Proceedings of Machine Learning Research},
  pages 370--378, Cadiz, Spain, 09--11 May 2016. PMLR.

\bibitem{deep_gaussian_processes}
Andreas Damianou and Neil~D. Lawrence.
\newblock Deep {G}aussian processes.
\newblock In Carlos~M. Carvalho and Pradeep Ravikumar, editors, {\em
  Proceedings of the Sixteenth International Conference on Artificial
  Intelligence and Statistics}, volume~31 of {\em Proceedings of Machine
  Learning Research}, pages 207--215, Scottsdale, Arizona, USA, 29 Apr--01 May
  2013. PMLR.

\bibitem{multi_layer_svm}
Marco~A. Wiering and Lambert~R.B. Schomaker.
\newblock {Multi-Layer} support vector machines.
\newblock In J.~A.~K. Suykens, M.~Signoretto, and A.~Argyriou, editors, {\em
  Regularization, optimization, kernels, and support vector machines}, pages
  457--476. Chapman and Hall/CRC, 2014.

\bibitem{suykensdeep2017}
Johan A.~K. Suykens.
\newblock Deep restricted kernel machines using conjugate feature duality.
\newblock {\em Neural Computation}, 29(8):2123--2163, August 2017.

\bibitem{salakhutdinov_restricted}
Ruslan Salakhutdinov, Andriy Mnih, and Geoffrey Hinton.
\newblock {Restricted Boltzmann Machines} for collaborative filtering.
\newblock In {\em {{ICML}} '07}, pages 791--798, Corvalis, Oregon, 2007. ACM
  Press.

\bibitem{sutskeverTemporalRestricted}
Ilya Sutskever and Geoffrey Hinton.
\newblock Learning multilevel distributed representations for high-dimensional
  sequences.
\newblock In Marina Meila and Xiaotong Shen, editors, {\em Proceedings of the
  Eleventh International Conference on Artificial Intelligence and Statistics},
  volume~2 of {\em Proceedings of Machine Learning Research}, pages 548--555,
  San Juan, Puerto Rico, 21--24 Mar 2007. PMLR.

\bibitem{sutskeverRecurrentTemporalRestricted}
Ilya Sutskever, Geoffrey~E Hinton, and Graham~W Taylor.
\newblock The recurrent temporal restricted boltzmann machine.
\newblock In {\em Advances in Neural Information Processing Systems},
  volume~21, 2008.

\bibitem{osogamiBoltzmannMachinesEnergybased2019}
Takayuki Osogami.
\newblock Boltzmann machines and energy-based models.
\newblock {\em arXiv:1708.06008 [cs]}, January 2019.

\bibitem{osogamiBoltzmannMachinesTimeseries2019}
Takayuki Osogami.
\newblock Boltzmann machines for time-series.
\newblock {\em arXiv:1708.06004 [cs]}, January 2019.

\bibitem{mercer_james_functions}
James Mercer.
\newblock Functions of positive and negative type, and their connection the
  theory of integral equations.
\newblock {\em Philosophical Transactions of the Royal Society of London.
  Series A, Containing Papers of a Mathematical or Physical Character},
  209(441-458):415--446, January 1909.

\bibitem{KRSL}
Y.~Engel, S.~Mannor, and R.~Meir.
\newblock The kernel recursive least-squares algorithm.
\newblock {\em IEEE Transactions on Signal Processing}, 52(8):2275--2285, 2004.

\bibitem{sw_krls}
S.~Van~Vaerenbergh, J.~Via, and I.~Santamaria.
\newblock A sliding-window kernel rls algorithm and its application to
  nonlinear channel identification.
\newblock In {\em 2006 IEEE International Conference on Acoustics Speech and
  Signal Processing Proceedings}, volume~5, pages V--V, 2006.

\bibitem{fd_krls}
Steven Van~Vaerenbergh, Ignacio Santamaría, Weifeng Liu, and José~C.
  Príncipe.
\newblock Fixed-budget kernel recursive least-squares.
\newblock In {\em 2010 IEEE International Conference on Acoustics, Speech and
  Signal Processing}, pages 1882--1885, 2010.

\bibitem{krls_tracker}
Steven Van~Vaerenbergh, Miguel Lázaro-Gredilla, and Ignacio Santamaria.
\newblock Kernel recursive least-squares tracker for time-varying regression.
\newblock {\em Neural Networks and Learning Systems, IEEE Transactions on},
  23:1313--1326, 08 2012.

\bibitem{SVM_timeseries}
K.~R. M{\"u}ller, A.~J. Smola, G.~R{\"a}tsch, B.~Sch{\"o}lkopf, J.~Kohlmorgen,
  and V.~Vapnik.
\newblock Predicting time series with support vector machines.
\newblock In Wulfram Gerstner, Alain Germond, Martin Hasler, and Jean-Daniel
  Nicoud, editors, {\em Artificial Neural Networks --- ICANN'97}, pages
  999--1004, Berlin, Heidelberg, 1997. Springer Berlin Heidelberg.

\bibitem{recurrent_lssvm}
Johan A.~K. Suykens and J.~Vandewalle.
\newblock Recurrent least squares support vector machines.
\newblock {\em IEEE Transactions on Circuits and Systems I: Fundamental Theory
  and Applications}, 47(7):1109--1114, 2000.

\bibitem{GP_regression}
Christopher Williams and Carl Rasmussen.
\newblock Gaussian processes for regression.
\newblock In D.~Touretzky, M.C. Mozer, and M.~Hasselmo, editors, {\em Advances
  in Neural Information Processing Systems}, volume~8. MIT Press, 1995.

\bibitem{sparse_gps}
Michalis Titsias.
\newblock Variational learning of inducing variables in sparse gaussian
  processes.
\newblock In David van Dyk and Max Welling, editors, {\em Proceedings of the
  Twelth International Conference on Artificial Intelligence and Statistics},
  volume~5 of {\em Proceedings of Machine Learning Research}, pages 567--574,
  Hilton Clearwater Beach Resort, Clearwater Beach, Florida USA, 16--18 Apr
  2009. PMLR.

\bibitem{deisenroth2015distributed}
Marc Deisenroth and Jun~Wei Ng.
\newblock Distributed gaussian processes.
\newblock In {\em International Conference on Machine Learning}, pages
  1481--1490. PMLR, 2015.

\bibitem{gp_lvm}
Neil Lawrence.
\newblock Probabilistic non-linear principal component analysis with gaussian
  process latent variable models.
\newblock {\em Journal of Machine Learning Research}, 6(60):1783--1816, 2005.

\bibitem{rbm_constrastive_divergence}
Geoffrey~E. Hinton.
\newblock Training products of experts by minimizing contrastive divergence.
\newblock {\em Neural Comput.}, 14(8):1771–1800, aug 2002.

\bibitem{salakhutdinov_deep}
Ruslan Salakhutdinov and Geoffrey Hinton.
\newblock {Deep Boltzmann Machines}.
\newblock In {\em proceedings of the Twelfth International Conference on
  Artificial Intelligence and Statistics}, volume 5 of JMLR, pages 448--455.
  PMLR, April 2009.

\bibitem{suykens2003support}
Johan A.~K. Suykens, Tony Van~Gestel, Joos Vandewalle, and Bart De~Moor.
\newblock A support vector machine formulation to {PCA} analysis and its kernel
  version.
\newblock {\em IEEE Transactions on Neural Networks}, 14(2):447--450, 2003.

\bibitem{HOUTHUYS202154}
Lynn Houthuys and Johan~A.K. Suykens.
\newblock Tensor-based restricted kernel machines for multi-view
  classification.
\newblock {\em Information Fusion}, 68:54--66, 2021.

\bibitem{TONIN2021661}
Francesco Tonin, Panagiotis Patrinos, and Johan~A.K. Suykens.
\newblock Unsupervised learning of disentangled representations in deep
  restricted kernel machines with orthogonality constraints.
\newblock {\em Neural Networks}, 142:661--679, 2021.

\bibitem{WinantLatentSpace}
David Winant, Joachim Schreurs, and Johan A.~K. Suykens.
\newblock Latent space exploration using generative kernel {PCA}.
\newblock In Bart Bogaerts, Gianluca Bontempi, Pierre Geurts, Nick Harley,
  Bertrand Lebichot, Tom Lenaerts, and Gilles Louppe, editors, {\em Artificial
  Intelligence and Machine Learning}, pages 70--82, Cham, 2020. Springer
  International Publishing.

\bibitem{ToninOOD}
Francesco Tonin, Arun Pandey, Panagiotis Patrinos, and Johan A.~K. Suykens.
\newblock Unsupervised energy-based out-of-distribution detection using
  stiefel-restricted kernel machine.
\newblock In {\em 2021 International Joint Conference on Neural Networks
  (IJCNN)}, pages 1--8, 2021.

\bibitem{joachim}
Joachim Schreurs and Johan A.~K. Suykens.
\newblock {Generative Kernel PCA}.
\newblock In {\em {{European Symposium on Artificial Neural Networks,
  Computational Intelligence and Machine Learning }}}, pages 129--134, 2018.

\bibitem{robust2020}
Arun Pandey, Joachim Schreurs, and Johan A.~K. Suykens.
\newblock Robust generative restricted kernel machines using weighted conjugate
  feature duality.
\newblock In {\em proceedings of the Sixth International Conference on Machine
  Learning, Optimization, and Data Science (LOD)}, 2020.

\bibitem{GENRKM}
Arun Pandey, Joachim Schreurs, and Johan~A.K. Suykens.
\newblock Generative restricted kernel machines: A framework for multi-view
  generation and disentangled feature learning.
\newblock {\em Neural Networks}, 135:177--191, 2021.

\bibitem{strkm}
Arun Pandey, Michaël Fanuel, Joachim Schreurs, and Johan A.~K. Suykens.
\newblock Disentangled representation learning and generation with manifold
  optimization.
\newblock {\em Neural Computation}, 34(10):2009--2036, 09 2022.

\bibitem{rrkm_esann}
Arun Pandey, Hannes De~Meulemeester, Henri De~Plaen, Bart De~Moor, and Johan
  Suykens.
\newblock Recurrent {Restricted} {Kernel} {Machines} for {Time}-series
  {Forecasting}.
\newblock In {\em {ESANN} 2022 proceedings}, pages 399--404, Bruges (Belgium)
  and online event, 2022. Ciaco - i6doc.com.

\bibitem{scholkopf1997kernel}
Bernhard Sch\"olkopf, Alexander Smola, and Klaus-Robert M\"uller.
\newblock Kernel principal component analysis.
\newblock In {\em International conference on artificial neural networks},
  pages 583--588. Springer, 1997.

\bibitem{mika_kernel_nodate}
Sebastian Mika, Bernhard Sch\"olkopf, Alex Smola, Klaus-Robert M\"uller,
  Matthias Scholz, and Gunnar R\"atsch.
\newblock Kernel pca and de-noising in feature spaces.
\newblock In {\em Advances in Neural Information Processing Systems}, pages
  536--542. MIT Press, 1999.

\bibitem{bui_projection-free_2019}
Anh~Tuan Bui, Joon-Ku Im, Daniel~W. Apley, and George~C. Runger.
\newblock Projection-free kernel principal component analysis for denoising.
\newblock {\em Neurocomputing}, 357:163--176, 2019.

\bibitem{kwok_pre-image_2004-2}
James~T. Kwok and Ivor Wai-Hung Tsang.
\newblock The pre-image problem in kernel methods.
\newblock {\em IEEE Transactions on Neural Networks}, 15:1517--1525, 2003.

\bibitem{honeine_preimage_2011-1}
Paul Honeine and Cedric Richard.
\newblock Preimage problem in kernel-based machine learning.
\newblock {\em IEEE Signal Processing Magazine}, 28(2):77--88, March 2011.

\bibitem{weston_learning_2004}
Jason Weston, Bernhard Sch\"{o}lkopf, and G\"{o}khan Bakir.
\newblock Learning to find pre-images.
\newblock In {\em Advances in Neural Information Processing Systems},
  volume~16, 2003.

\bibitem{santafe}
A.S. Weigend and N.A. Gershenfeld.
\newblock {\em Time Series Prediction: {F}orecasting the Future and
  Understanding the Past}.
\newblock Addison-Wesley, 1993.

\bibitem{chickenpox}
Benedek Rozemberczki, Paul Scherer, Oliver Kiss, Rik Sarkar, and Tamas Ferenci.
\newblock Chickenpox cases in hungary: a benchmark dataset for spatiotemporal
  signal processing with graph neural networks, 2021.

\bibitem{appliances_energy}
Luis~M. Candanedo, Véronique Feldheim, and Dominique Deramaix.
\newblock Data driven prediction models of energy use of appliances in a
  low-energy house.
\newblock {\em Energy and Buildings}, 140:81--97, 2017.

\bibitem{gas_turbine}
Heysem Kaya, Pınar Tüfekci, and Erdinç Uzun.
\newblock Predicting co and nox emissions from gas turbines: Novel data and a
  benchmark pems.
\newblock {\em Turkish Journal of Electrical Engineering and Computer
  Sciences}, 27(6):4783--4796, 11 2019.

\end{thebibliography}

\end{document}